\documentclass[final,authoryear,5p,times,twocolumn]{elsarticle}

\usepackage{bbm}
\usepackage[english]{babel}
\usepackage[utf8]{inputenc} 
\usepackage[T1]{fontenc}    
\usepackage{hyperref}       
\usepackage{url}            
\usepackage{booktabs}       
\usepackage{amsfonts}       
\usepackage{nicefrac}       
\usepackage{microtype}      
\usepackage{algorithm}
\usepackage{algorithmic}
\usepackage{graphicx}
\usepackage{subfigure}
\usepackage{amsmath}
\usepackage{amsthm}
\usepackage{color}
\usepackage{multirow}
\usepackage{mathrsfs}
\usepackage{float}
\usepackage[misc]{ifsym}
\usepackage{caption}
\usepackage{amsthm}
\usepackage{graphicx}
\usepackage{pythonhighlight}
\usepackage{bm}



\journal{Neural Networks}

\newtheorem{definition}{Definition}[section]

\newtheorem{theorem}{Theorem}[section]
\newtheorem{assumption}{Assumption}[section]

\begin{document}

\title{Information Theory-Guided Heuristic Progressive Multi-View Coding}

\begin{frontmatter}


\author{Jiangmeng Li\fnref{label1,label2}}
\address[label1]{Science \& Technology on Integrated Information System Laboratory, Institute of Software, Chinese Academy of Sciences, Beijing, China. \fnref{label1}}
\address[label2]{University of Chinese Academy of Sciences, Beijing, China. \fnref{label2}}
\author{Hang Gao\fnref{label1,label2}}
\author{\Letter Wenwen Qiang\fnref{label1,label2}}
\ead{wenwen2018@iscas.ac.cn}
\author{Changwen Zheng\fnref{label1}}

\begin{abstract}
Multi-view representation learning aims to capture comprehensive information from multiple views of a shared context. Recent works intuitively apply contrastive learning to different views in a pairwise manner, which is still scalable: view-specific noise is not filtered in learning view-shared representations; the \textit{fake} negative pairs, where the negative terms are actually within the same class as the positive, and the \textit{real} negative pairs are coequally treated; evenly measuring the similarities between terms might interfere with optimization. Importantly, few works study the theoretical framework of generalized self-supervised multi-view learning, especially for more than two views. To this end, we rethink the existing multi-view learning paradigm from the perspective of information theory and then propose a novel information theoretical framework for generalized multi-view learning. Guided by it, we build a multi-view coding method with a three-tier progressive architecture, namely \textit{Information theory-guided heuristic Progressive Multi-view Coding} (IPMC). In the \textit{distribution-tier}, IPMC aligns the distribution between views to reduce view-specific noise. In the \textit{set-tier}, IPMC constructs self-adjusted contrasting pools, which are adaptively modified by a view filter. Lastly, in the \textit{instance-tier}, we adopt a designed unified loss to learn representations and reduce the gradient interference. Theoretically and empirically, we demonstrate the superiority of IPMC over state-of-the-art methods.
\end{abstract}


\end{frontmatter}

\section{Introduction}\label{sec:introduction}

One of the fundamental ideas behind self-supervised learning (SSL) lies in designing appropriate self-supervised objectives without manual annotations. Recent works explore how to employ the maximization of the mutual information (MI) between the inputs and outputs of the encoder to learn discriminative representations from a single view \cite{belghazi2018, hjelm2018learning}. Yet a single view may not provide sufficient information and data is usually observed by individuals through multiple views. Multi-view learning (MVL) \cite{2016Multikan, ZhangHcFZH19, Tian2019Contrastive} therefore aims at capturing information shared among views to enhance multi-view representations. 

Existing self-supervised MVL methods perform anchor-based pairwise (e.g., an anchor term and a positive term form a pair) contrastive learning (CL) among views through adopting sophisticated data augmentation and specific encoders \cite{2020DataHenaff, 2018RepresentationOord, Tian2019Contrastive, 2019Philip}. However, such a pairwise-based paradigm suffers from the disturbance caused by view-specific noise, the misallocated fake negative terms, and the optimization instability on account of the undifferentiated measurement of the similarities, which jointly make the state-of-the-art self-supervised MVL unable to fully model the conjunction information from multiple views. \cite{2020Hard} aims to ease the issue caused by misallocated fake negative terms by proposing a sampling method to get hard negative samples, but the intrinsic issue can not be addressed. 

{From the perspective of information theory, benchmark methods \cite{2020Tsai, li2022modeling} revisit the learning paradigm of conventional self-supervised MVL methods, and further propose the corresponding solutions, which enlightens the researchers and demonstrates the importance of adopting the information theory to analyze the self-supervised MVL methods. The intrinsic intuition behind the importance of using the information theory is that such a theory can sufficiently describe the interpretability of self-supervised MVL methods, and the long-lasting issues of the methods can also be well demonstrated, such that the information theory is an appropriate guide to researchers. However, there only exist limited self-supervised MVL research dedicated to tackle the mentioned issues guided by an integrated theoretical framework, and the issues of the disturbance caused by view-specific noise, the misallocated fake negative terms, and the optimization instability on account of the undifferentiated measurement of the similarities are still not well analyzed from the perspective of information theory.}
 
In this paper, we first propose a comprehensive information theory-based framework for generalized multi-view learning. Guided by it, we rethink the mentioned issues of pair-wise learning paradigm and figure out their influences on conventional CL: 1) the view-specific noise causes the inconsistency of the learned representations; 2) the misallocated fake negative terms may lead the learning process under biased self-supervision, as a consequence, the learned representations capture wrong discriminative information; and 3) the optimization instability because of evenly measuring the similarities causes the insufficiency of self-supervision and the increase of the training complexity.

 \begin{figure*}
	\centering
	\includegraphics[width=0.82\textwidth]{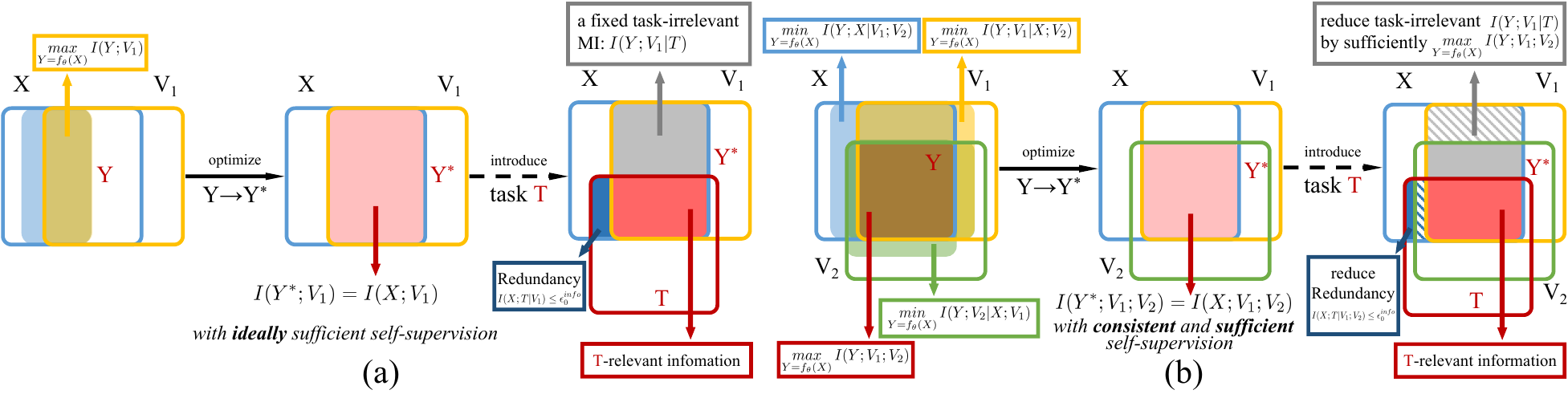}
	\caption{Different from the conventional self-supervised MVL paradigm shown in (a), the proposed generalized self-supervised MVL paradigm learns consistent and sufficient representations $Y^*$ from more than two views, which can naturally reduce the redundancy $\epsilon^{info}_i$ and task-irrelevant information $I(Y;V_i|T)$, where $i \in \{1, ..., m\}$. As an example shown in (b), guided by it, IPMC adopts a three-tier progressive architecture to learn consistent and sufficient representations by maximizing $I(Y;V_1;V_2)$ and minimizing $I(Y;X|V_1;V_2)$, $I(Y;V_1|X;V_2)$, and $I(Y;V_2|X;V_1)$.}
	\label{fig:theoretical}
	\vspace{0.5cm}
\end{figure*}

Our goal is to learn consistent and sufficient representations for generalized multi-view self-supervision. To this end, we develop a self-supervised MVL approach, called IPMC, under the guidance of the proposed framework, which can robustly capture view-shared information through a three-tier progressive architecture. In the \textit{distribution-tier}, we are concerned that different views of the same sample reflect the same semantic content, but the learned representations may differ greatly for different views and contain irrelevant information for distinguishing the views rather than the semantic contents. Therefore, we directly align the distributions of different views by minimizing a discrepancy metric between them, e.g., KL-divergence \cite{2003Goldbberger} and Wasserstein distance \cite{2017Mart, 2019Kuroki}, to capture view-shared semantic information and discard task-irrelevant view-dependent noise. In the \textit{set-tier}, IPMC innovatively waives the \textit{anchor-based} pairwise contrast and instead utilizes self-adjusted pool-based contrast. The self-adjusted pool can be dynamically modified by using a designed view filter to transfer fake negative terms to the positive pool. Next, in the \textit{instance-tier}, motivated by \cite{2018ArcFace, Circle2}, we adopt a unified loss function to improve IPMC by emphasizing the weight of similarities that have larger contributions to the optimization \cite{2016LiuWYY, 2017LiuWYLRS, 2018AdditiveWang, 2018CosFaceWang}. As a result, IPMC can boost the lower bound of the MI between views.

The major contributions of this paper include: 1) we reformulate the MVL paradigm from the perspective of information theory and propose a generalized self-supervised MVL framework (especially for more than two views); 2) guided by the proposed information theoretical framework, we introduce a novel multi-view coding method, called IPMC, which can efficiently learn consistent and sufficient representations; 3) we provide both theoretical analysis and extensive empirical evaluations to show the superiority of IPMC.

\section{Related works}
\subsection{Unsupervised learning}
Classic unsupervised learning methods \cite{Rao-et-al:nonmon} learn the latent manifold of unlabeled data \cite{bengio2013representation}. SSL methods \cite{li2022metaug, qiang2022interventional} capture useful information from unlabeled data by constructing auxiliary tasks, where the supervised information is automatically constructed from the data to train deep neural networks. Self-encoding approaches \cite{hinton2006reducing, Kingma2014Auto, alemi2016deep}, as a category of SSL, learn an encoder network to extract representations holding the principle of the coding theory\cite{2013Yoshua}. Plenty of the adversarial methods \cite{Generative1, makhzani2015adversarial, 2016Oord, donahue2016adversarial, Splitp2} are also based on the coding theory, which estimates generative models via an adversarial process. 
Efforts have also been made to explore SSL approaches in specific fields \cite{2010Vincent} such as NLP and computer vision (CV) \cite{2018Devlin, 2018Semanet, qiang2023meta}. CL-based SSL methods such as NAT \cite{bojanowski2017unsupervised} and DIM \cite{hjelm2018learning} have achieved state-of-the-art performances in CV.

\begin{figure*}
	\centering
	\includegraphics[width=0.82\textwidth]{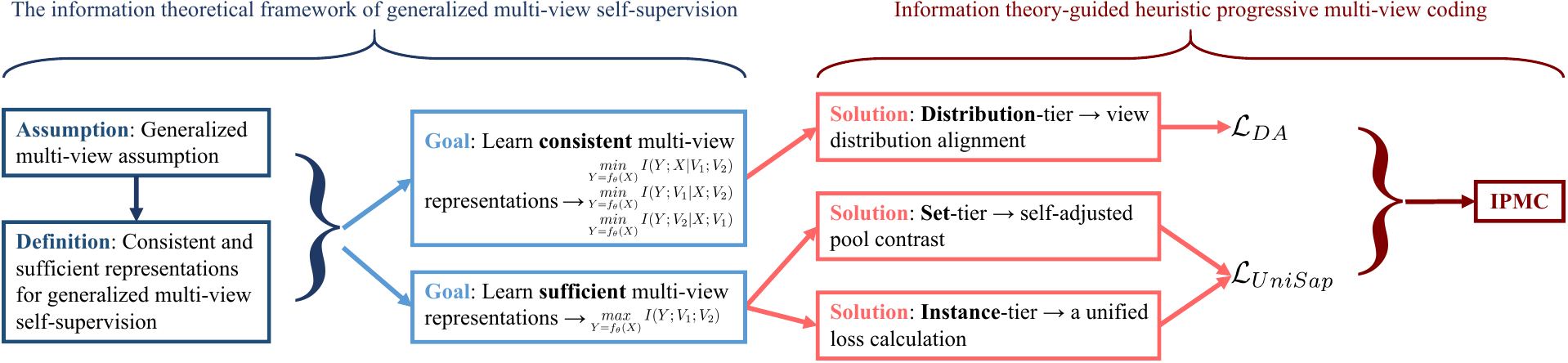}
	\caption{The pipeline of our proposed information theoretical framework and the corresponding coding method, i.e., IPMC.}
	\label{fig:algoitmatch}
	\vspace{0.5cm}
\end{figure*}

\subsection{Multi-view learning}
As demonstrated in \cite{2013A2}, MVL methods achieved impressive success in various fields \cite{Shiliang2011Multi, 2010An}. The fundamental idea of MVL is to obtain useful features by considering information from multiple views and then find the relationship between them (e.g., the complementary relationship, the consistent relationship, etc.). Currently, a popular learning paradigm in this field is cross-view learning, which searches mappings between two views in an encoder-decoder manner and has been widely applied in practical applications \cite{2010Rasiwasia, 2016LearningCastrejon, 2018UnsupervisedChung}. Recent works explore self-supervised MVL methods, e.g., CPC \cite{2018RepresentationOord}, AMDIM \cite{2019Philip}, SwAV \cite{2020Mathilde}, CMC \cite{Tian2019Contrastive}, MoCo \cite{2020Kaiming}, SimCLR \cite{chen2020simple}, DebiasedCL\cite{2020Debiased}, HardCL \cite{2020Hard}, BYOL \cite{2020Bootstrap}, and Barlow Twins \cite{2021Barlow}. These methods learn representations by capturing information shared among multiple views. However, these self-supervised MVL methods generally follow the pairwise CL paradigm and focus on constraining the distance of features in the set level or instance level, while the feature distributions of different views in the latent space are not explicitly considered.

Additionally, only a few works\cite{2008Sridharan, 2020Tsai} provide solid theoretical analyses about self-supervised MVL, but these theories still have limitations or drawbacks for analyzing learning representations from more than two views.

\subsection{Distribution alignment}
{Aligning distributions of different domains was widely applied in transfer learning, which largely enhance the performance for transfer learning approaches, and thus, transfer learning methods achieve significant successes in various application fields \cite{fateh2021multilingual}. Distribution alignment methods can be divided into three categories \cite{2021Fuzhen}, i.e., the instance-based approach \cite{2011Chen}, the parameter-based approach \cite{2019You}, and the representation learning (RL) based approach \cite{2019Wu, 2019Zhao}. Many RL-based approaches align the distributions by minimizing a certain metric \cite{2017Cariucci, 2018Sohn}, including KL-divergence, and Wasserstein distance \cite{2017Mart, 2019Kuroki, qiang2021robust, qiang2021auxiliary}, etc. In this paper, we apply distribution alignment to MVL for constraining the representations from the global distribution tier.}

\section{Theoretical framework}
\label{sec:method}
{\bf{Notation.}} Fig. \ref{fig:theoretical} illustrates our proposed generalized self-supervised MVL paradigm and IPMC by using information theoretical description. We regard the input random variable $X$ (may be considered as $V_0$) and the other self-supervised signals (e.g., $V_1$ and $V_2$) as the views of the original data. $Y$ is the representation learned by a deterministic encoder $f_{\theta}(\cdot)$, i.e., $Y=f_{\theta}(X)$. $T$ denotes the desired task-relevant information. For random variables $A$, $B$, and $C$, $H(A)$ denotes the entropy of $A$, and $I(A;B)$ represents the MI of $A$ and $B$. Accordingly, $H(A|B)$ denotes the conditional entropy of $H(A)-I(A;B)$, and $I(A;B|C)$ represents the conditional MI of $A$ and $B$ given $C$. $I(A;B|C;D)$ denotes the conditional MI between the two random variables $A$ and $B$ given $C$ and $D$.

To clarify the terminology, we define \textit{sample} as a multi-view input image, and \textit{view} denotes the macroscopic definition of a view (e.g., RGB, L, ab views), and \textit{term} denotes the microcosmic definition of a view of a specific image.

\subsection{Generalized multi-view assumption for more than two views}
To derive the information theory-based diagram, we extend the common two-view assumption \cite{2008Sridharan, 2013Xu}. The introduced assumption can generally describe the self-supervised MVL among multiple (especially more than two) self-supervised signals:
\begin{assumption}
	\label{ass:1}
	The $m$ self-supervised signals ($V$) are approximately redundant to the input for the task-relevant information. Namely, there exist a set $\{\epsilon^{info}_i>0\}$, where $i \in \{1,...,m\}$, such that, for each $\epsilon^{info}_i$, we have $I(X;T|V_i) \leq \epsilon^{info}_i$.
\end{assumption}
Assumption \ref{ass:1} states that, for each $\epsilon^{info}_i$, when it is small, the task-relevant information mainly lies in the MI between the input and the self-supervised signal. For more than two views, when $m$ is not large, as $m$ increases, $I(X;T|\{V_i\}_{i=1}^{m})$ gets smaller, since the quantity of constraints $\{\epsilon^{info}_i>0\}_{i=1}^{m}$ of the MI is also growing, and then the task-relevant discriminative information is more likely to lie in the MI between views, i.e., $I(X;\{V_i\}_{i=1}^{m})$. It is supported by the view-vanishing experiments (See App. \ref{sec:extendedcomp}). Therefore, compared with the Multi-view assumption of \cite{2020Tsai}, our proposed generalized multi-view assumption better depicts the improvement of introducing more than two views. Note that the downstream task $T$ can be classification, clustering, regression, etc.

To learn discriminative representation $Y$ from $X$, classic works \cite{hjelm2018learning, 1999Tishby, 2017Achille, 2020Tsai} maximize $I(Y;V_1)$ with a single self-supervised signal. Based on the conventional multi-view assumption, the prior information bottleneck methods \cite{1999Tishby, 2017Achille} minimize $I(Y;X)$ to reduce the complexity of $Y$ and discard task-irrelevant information. Inspired by that, in \cite{2020Tsai}, $H(Y|V_1)$ is further minimized by adopting an information bottleneck-based method. Yet, as demonstrated in Fig. \ref{fig:theoretical} (a), this learning paradigm still faces three intractable issues: 1) there is a \textit{fixed} task-irrelevant MI, i.e., $I(Y;V_1|T)$, which cannot be reduced; 2) the redundancy, i.e., $I(X;T|V_1) \le \epsilon^{info}_1$, is hard to reduce, which results in less task-relevant information in the MI between views, thus undermining the discriminability of the learned representation ($Y^*$); and 3) the conditional entropy $H(Y|X)$ is not reduced.

Fortunately, under the proposed Generalized multi-view assumption, we find that introducing more views can well solve the mentioned issues: 1) the fixed task-irrelevant MI $I(Y;V_1|T)$ may also be reduced by jointly maximizing $I(Y;V_1;V_2)$. By the same token, the task-irrelevant MI caused by learning from $X$ and $V_2$ can also be reduced by introducing $V_1$; 2) the redundancy $I(X;T|V_1) \le \epsilon^{info}_1$ generated by learning from $X$ and a single self-supervised signal $V_1$ can be alleviated by adopting more views, and specifically, the reduced redundancy is $I(X;T|V_1;V_2) \le \epsilon^{info}_1$; and 3) more views can improve the potential to reduce the view-specific information.

\begin{figure*}
	\centering
	\includegraphics[width=0.82\textwidth]{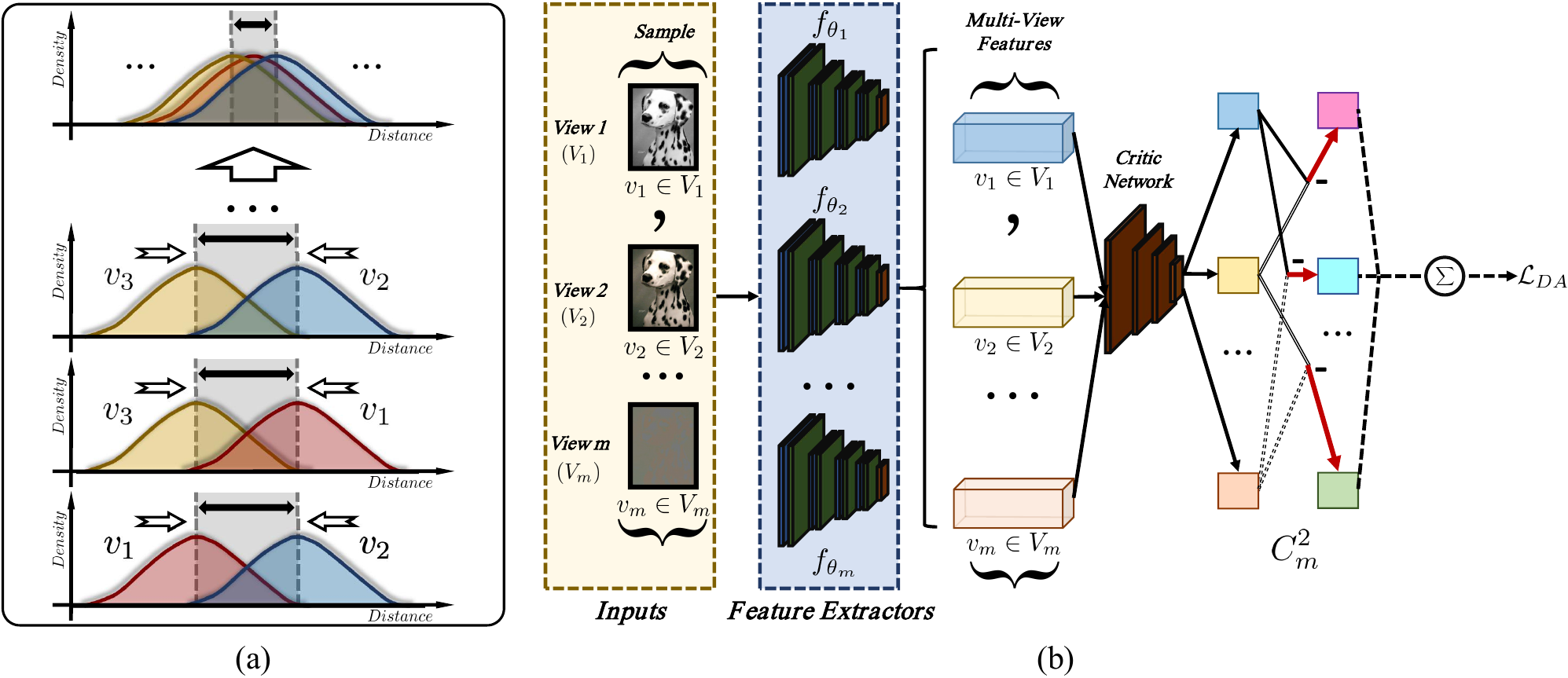}
	\caption{(a) The conceptual illustration of view distribution alignment. (b)  The detailed architecture. IPMC aligns the learned features of $m$ views from $n$ samples by minimizing the discrepancy metric to reduce the view-specific noise.}
	\label{fig:alignment}
	\vspace{0.5cm}
\end{figure*}

\subsection{Learning consistent and sufficient representations from multiple views}
Based on the proposed generalized multi-view assumption, as Assumption \ref{ass:1}, our goal is to learn discriminative representations from multiple views (more than two views) with consistent and sufficient self-supervision.

However, introducing more views could pose new issues, for instance: 1) vast view-dependent noisy information is brought in MVL, which causes inconsistency of the learned representations; 2) conventional CL does not distinguish the \textit{fake} negative pairs, where the negative terms are actually within the same class as the positive, from the \textit{real} negative pairs, thus may undermine the sufficiency of self-supervision, and the representations may incorporate wrong discriminative features, which causes that the self-supervision is biased; and 3) evenly measuring the similarities between terms might interfere with optimization, cause self-supervision insufficiency, and increase the training complexity. In consideration of these, we propose consistent and sufficient representations for generalized multi-view self-supervision as follows:
\begin{definition}
	\label{def:1}
	Consistent and sufficient representations: Suppose $Y^*$ denotes the multi-view representation learned from more than two views. $Y^*$ is the consistent and sufficient representation if and only if: $Y^*=\mathop{\text{argmin}} \limits_{Y}I(Y;X|V_1;V_2)$, $Y^*=\mathop{\text{argmin}} \limits_{Y}I(Y;V_1|X;V_2)$, $Y^*=\mathop{\text{argmin}} \limits_{Y}I(Y;V_2|X;V_1)$, and $Y^*=\mathop{\text{argmax}} \limits_{Y}I(Y;V_1;V_2)$ jointly hold.
\end{definition}
Definition \ref{def:1} defines our proposed generalized multi-view self-supervised representation learning paradigm. Under this diagram, we can jointly reduce the redundancy $\epsilon^{info}_i$ and task-irrelevant $I(Y;V_i|T)$, for $i \in \{1,...,m\}$, to learn discriminative representations by utilizing more than two views. As demonstrated in Fig. \ref{fig:algoitmatch}, the constraint of consistency, i.e., jointly minimizing $I(Y;X|V_1;V_2)$, $I(Y;V_1|X;V_2)$, and $I(Y;V_2|X;V_1)$, globally requires the representations to learn view-shared information $I(X;V_1) + I(X;V_2) + I(V_1;V_2)$. Under this constraint, performing sufficient self-supervision, i.e., maximizing $I(Y;V_1;V_2)$, is more achievable. Note that it is straightforward to generalize Definition \ref{def:1} to more than three views.

\section{Method description}
To achieve the desired multi-view representations, as defined in \ref{def:1}, we propose an information theory-guided heuristic progressive multi-view coding method, called IPMC, which is realized as a three-tier learning architecture. As shown in Fig. \ref{fig:algoitmatch}, IPMC minimizes $I(Y;X|V_1;V_2)$, $I(Y;V_1|X;V_2)$, and $I(Y;V_2|X;V_1)$ by view distribution alignment to reduce the view-dependent noises in the distribution-tier. To maximize the MI $I(Y;V_1;V_2)$ and then acquire sufficiently self-supervised representations, IPMC utilizes a designed self-adjusted pool contrast in the set-tier and a unified loss in the instance-tier. Implementation details of IPMC are presented in Sec. \ref{sec:imageclass}.

Formally, we consider the multi-view dataset ${X^m} = \left[ {x_1^m,x_2^m,...,x_n^m} \right]$, where $X^m$ represents the sample collection from the $m$-th view, and ${x_i^m}$, $i \in \{1,...,n\}$ represents the $m$-th view of $i$-th sample. $n$ is the number of samples. We denote $X$ as a variable that is sampled \textit{i.i.d} from distribution $\mathcal{P}\left(X \right) $. Also, we denote $X^m$ as a random variable sampled \textit{i.i.d} from the distribution $\mathcal{P}\left(X^m \right) $. We denote the similarity of homogeneous features as $\bm{S^{pos}}$, and the similarity of heterogeneous features is denoted as $\bm{S^{neg}}$.

\begin{figure*}
	\centering
	\includegraphics[width=0.82\textwidth]{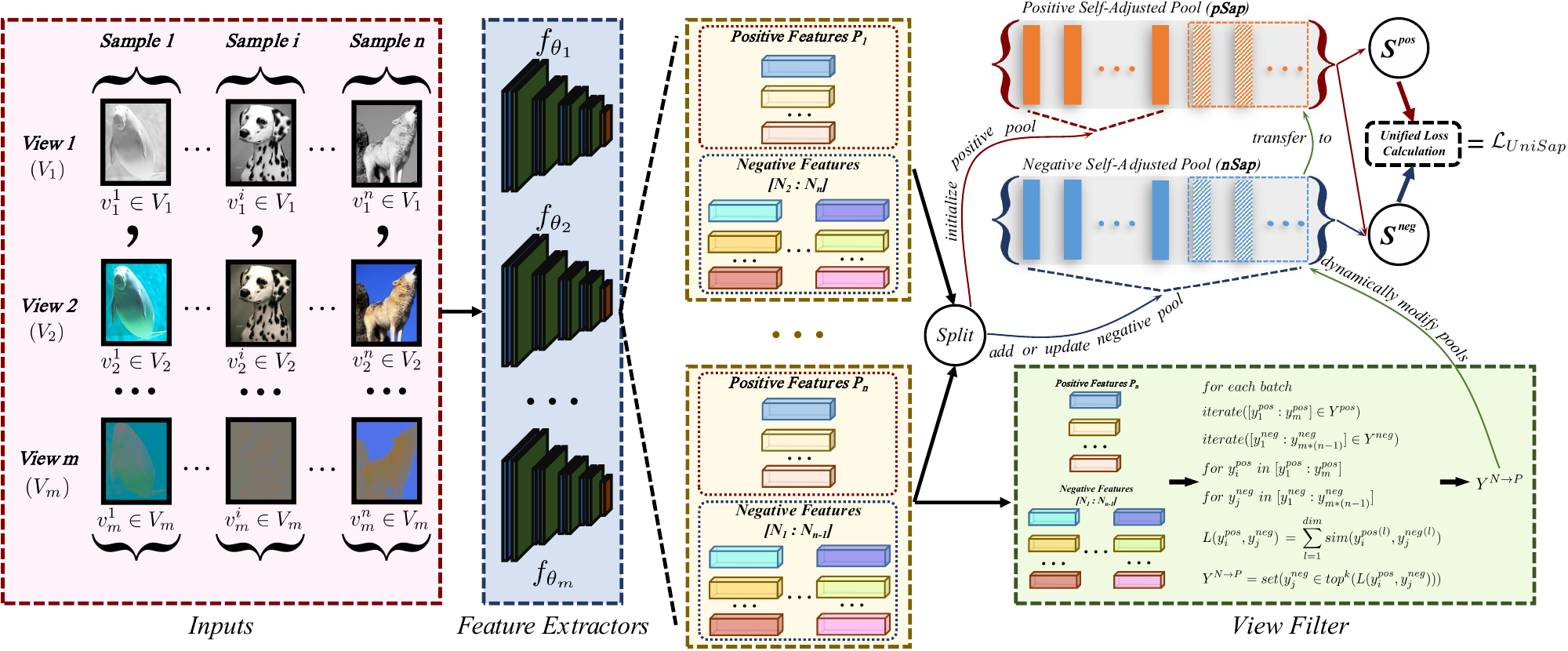}
	\caption{Our proposed self-adjusted pool contrast utilizes a designed view filter to pick out fake negative terms from the negative pool, and then accordingly transfers them to the positive pool.}
	\label{fig:algoframe}
	\vspace{0.5cm}
\end{figure*}

\subsection{Distribution-tier: view distribution alignment}
We harbor the foundational idea that consistent multi-view representations are more discriminative, since MVL focuses on comprehensively capturing the crucial and discriminative information that is shared among different views.
\begin{theorem}
	\label{the:1}
	Suppose $Y$ is the inconsistent representation with the redundant view-specific information $I(Y;X|V_1;V_2)$, $I(Y;V_1|X;V_2)$, and $I(Y;V_2|X;V_1)$, there exists a $Y^*$ that is the consistent representation with the minimized view-specific information $I^{min}(Y;X|V_1;V_2)$, $I^{min}(Y;V_1|X;V_2)$, and $I^{min}(Y;V_2|X;V_1)$ s.t. $H(Y^*) = I(X;V_1;V_2) + I^{min}(Y;X|V_1;V_2) + I^{min}(Y;V_1|X;V_2) + I^{min}(Y;V_2|X;V_1) \leq H(Y) = I(X;V_1;V_2) + I(Y;X|V_1;V_2) + I(Y;V_1|X;V_2) + I(Y;V_2|X;V_1)$ so that the consistency constraint can improve the compactness of the learned representation.
\end{theorem}
{See App. \ref{sec:proof} for the proof of validating that there exists a $Y^*$ s.t. $H(Y^*) \leq H(Y)$. Based on Theorem \ref{the:1}, to learn consistent and compressed representations by discarding view-specific noise, we propose to minimize $I(Y;X|V_1;V_2)$, $I(Y;V_1|X;V_2)$, and $I(Y;V_2|X;V_1)$ in the distribution-tier. We align the distributions of views, i.e., $\mathcal{P}\left(X^1 \right) $, $\mathcal{P}\left(X^2 \right)$,...,$\mathcal{P}\left(X^m \right)$, by minimizing a specific discrepancy metric, i.e., Wasserstein distance, between them. See Fig. \ref{fig:alignment} for details.}

For $\mathcal{P}\left(X^i \right) $ and $\mathcal{P}\left(X^j \right) $, where $i,j \in \left\{ {1,...,m}\right\} \cap i \neq j$, the $p{\rm{th}}$ Wasserstein distance-based discrepancy metric can be calculated as:
\begin{equation}
	\begin{array}{l}
		{W_p}\left( {\mathcal{P}\left(X^i \right),\mathcal{P}\left(X^j \right)} \right) \\ =
		{\left( {\mathop {\inf }\limits_{\mu \left( {x^i,x^j} \right) \in \Pi \left( {x^i,x^j} \right)} \int {c{{\left( {x^i,x^j} \right)}^p}d\mu } } \right)^{\frac{1}{p}}},
	\end{array}
\end{equation}
where $p \in \left\{{1,...,C_m^2}\right\}$, and $C_m^2$ denotes the number of combination of views. $c\left( {x^i,x^j} \right)$ represents the distance of two patterns, and $\Pi \left( {x^i,x^j} \right)$ denotes the set of all joint distributions $\mu \left( {x^i,x^j} \right)$ that satisfy $\mathcal{P}\left(X^i \right) = \int_x^j {\mu \left( {x^i,x^j} \right)dx^j} ,{\mathcal{P}\left(X^j \right)} = \int_x^i {\mu \left( {x^i,x^j} \right)dx^i}$.

Based on the Kantorovich-Rubinstein theorem, the dual form of Wasserstein distance can be written as:
\begin{equation} \label{eq:wdcon}
	\begin{array}{l}
		{W_p}\left( {\mathcal{P}\left( {{X^i}} \right),\mathcal{P}\left( {{X^j}} \right)} \right)\\
		= \mathop {\sup }\limits_{\left\| \gamma \right\|{}_L \le 1} \mathop E\limits_{{x^i} \sim \mathcal{P}\left( {{X^i}} \right)} \left[ {\gamma \left( {{x^i}} \right)} \right] - \mathop E\limits_{{x^j} \sim \mathcal{P}\left( {{X^j}} \right)} \left[ {\gamma \left( {{x^j}} \right)} \right],
	\end{array}
\end{equation}
where $\gamma :{x} \to R$ is the 1-Lipschitz function and satisfies ${\left\| \gamma \right\|_L} = \mathop {\sup }\limits_{x \ne y} {{\left| {\gamma \left( x \right) - \gamma \left( y \right)} \right|} \mathord{\left/
		{\vphantom {{\left| {\gamma \left( x \right) - \gamma \left( y \right)} \right|} {\left| {x - y} \right|}}} \right.
		\kern-\nulldelimiterspace} {\left| {x - y} \right|}} \le 1$.

{Theoretically, many divergences can be generalized as the discrepancy metric, but we adopt the Wasserstein distance in IPMC because it has an outstanding gradient superiority in this task compared with other discrepancy metrics, e.g., KL-divergence, H-divergence, etc.}

{For ease of description, we elaborate on the intrinsic behavior of the view distribution alignment and further perform analysis. To align the distributions of views, we map data into a latent space to learn representations by using a neural network, and then measure the distance based on the discrepancy metric. The distribution of representations usually exist throughout the latent space, since the mapping network reduces the dimensionality of representations. For the conventional discrepancy metric, e.g., KL-divergence, the latent features of samples in a region where the probability of a certain distribution is extremely greater than other distributions have little contribution to the gradient with the contrastive loss \cite{2022Chaos}. Yet, the gradient, computed by using Wasserstein distance, maintains its consistency for different sample latent features. If sample features are indistinguishable based on a discrepancy metric, the gradient vanishing problem would be unable to be eliminated, since the distributions have supports lying on low dimensional manifolds in the latent space \cite{2010Narayanan, 2017Mart}. Compared with conventional discrepancy metrics, adopting Wasserstein distance can avoid such a case to a significant extent.}

{To further understand the gradient superiority of Wasserstein distance, we provide a practical derivation on the gradients of Wasserstein distance. Specifically, to fit the practical data, we transform Eq. \ref{eq:wdcon} into the discrete analog form \cite{DBLP:conf/icml/DuklerLLM19}, and then derive the corresponding gradient of Wasserstein distance by}
\begin{equation} \label{eq:wddis}
	\begin{array}{l}
		\textrm{DiscGrad}\left[{W_p}\left( {\mathcal{P}\left( {{X^i}} \right),\mathcal{P}\left( {{X^j}} \right)} \right)\right]\\
		= \mathop {\sum }\limits_{{x^i} \sim \mathcal{P}\left( {{X^i}} \right), \ {x^j} \sim \mathcal{P}\left( {{X^j}} \right)} \left({\gamma \left( {{x^i}} \right)} - {\gamma \left( {{x^j}} \right)} \right)^2 \cdot \frac{x^i / di + x^j / dj}{2},
	\end{array}
\end{equation}
{where $\textrm{DiscGrad}\left[\cdot\right]$ denotes the function computing the gradient of the specific discrete analog form. The intuition behind such a behavior is that the intrinsic idea behind the view distribution alignment is to appropriately reduce the domain shift between views, such that the \textit{exact} values of Wasserstein distances are not necessary. From Eq. \ref{eq:wddis}, we observe that compared with conventional discrepancy metrics, Wasserstein distance can provide more consistent gradients for each feature due to the ingredients of its functions, e.g., non-normalization. Such a theoretical conclusion is further supported by the empirical experiments in Sec. \ref{sec:deepexplore}, where we conduct experiments to compare the performance of using different discrepancy metrics to demonstrate the superiority of adopting Wasserstein distances.}

{Then, the ultimate loss of the view distribution alignment is defined as:}
\begin{equation}
	\mathcal{L}_{DA} = \sum\limits_{i,j \in \left\{ {1,...,m}\right\} \cap i \neq j} {W_p}\left( {\mathcal{P}\left( {{X^i}} \right),\mathcal{P}\left( {{X^j}} \right)} \right).
\end{equation}

{By utilizing the proposed view distribution alignment, IPMC can minimize $I(Y;X|V_1;V_2)$, $I(Y;V_1|X;V_2)$, and $I(Y;V_2|X;V_1)$ to acquire consistent multi-view representations from more than two views.}

\subsection{Set-tier: self-adjusted pool contrast}
Based on Definition \ref{def:1}, we jointly utilize the self-adjusted pool contrast in the set-tier and a unified loss in the instance-tier to learn sufficient representations.

As manifested in Fig. \ref{fig:algoframe}, in the set-tier, the proposed self-adjusted pool contrast groups the alternative terms into two dynamic pools. We separately use the encoders to embed features from different views. Initially, the features $\bm{Y}^{pos}=\lbrace {y}_{i}^{pos} \rbrace_{i=1}^{m}$ are extracted from the positive terms, and accordingly $\bm{Y}^{neg}=\lbrace {y}_{j}^{neg} \rbrace_{j=1}^{m*(n-1)}$ denotes the negative terms. Then, $\bm{Y}^{pos}$ and $\bm{Y}^{neg}$ are separately collected as the primary positive and negative pools. Yet there is a nonnegligible issue within the initialized pools: the primary positive pool cannot include all real positive terms, and the negative pool contains certain fake negative terms which are from the same class as the positive terms.

{To this end, we propose to adjust the pools to dynamically correct the fake negatives. That is, after a few starting epochs, we iteratively measure the similarities between each positive term and the negative terms within a batch and transfer top-$k$ similar negative terms to the positive pool. Behind this self-adjusted technique, we follow an essential conjecture that when the encoders have been trained by a few epochs, we can pick out the desired $real$ positive terms using a designed $k$-nearest neighbor method, i.e., $Y^{N \to P} = set({y}_{j}^{neg} \in top^{k}(L({y}_{i}^{pos}, {y}_{j}^{neg})))$, where ${y}_{i}^{pos}$ denotes the terms from positive pool and ${y}_{j}^{neg}$ denotes the terms from negative pool, respectively. The similarity is calculated by $cos({y}_{i}^{pos}, {y}_{j}^{neg}) = \frac{{{y}_{i}^{pos}} \times {{y}_{j}^{neg}}}{||{{y}_{i}^{pos}}|| \times ||{{y}_{j}^{neg}}||}$.}

The number of $\bm{Y}^{pos}$ is still exceedingly less than that of $\bm{Y}^{neg}$. To enrich $\bm{Y}^{pos}$, and appropriately increase the difficulty of the self-supervised task, we apply the memory bank \cite{un2}, which is self-updated with the calculation results in each training step. When current $\bm{Y}^{pos}$ is obtained, we also extract the past $\bm{Y}^{pos}$ records from the memory bank, which work as additional features of the positive terms and are added to $\bm{Y}^{pos}$. As for $\bm{Y}^{neg}$, instead of calculating them every time, we directly derive the past features from the memory bank.
\begin{figure}
	\centering
	\includegraphics[width=0.48\textwidth]{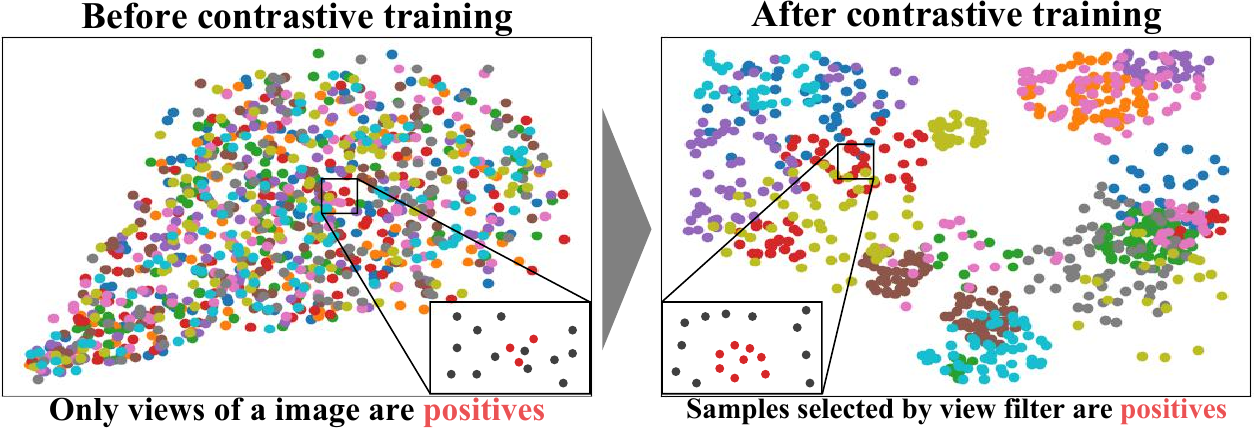}
	\caption{t-SNE visualization on CIFAR-10 with \textit{conv} encoder shows the latent clustered features learned by IPMC.}
	\label{fig:cltrainplot}
	\vspace{0.5cm}
\end{figure}
Then $\bm{S^{pos}}$ and $\bm{S^{neg}}$ are calculated from $\bm{Y}^{pos}$ and $\bm{Y}^{neg}$. We suppose that the modified $\bm{Y}^{pos}$ contains $n^{pos}$ terms, and $\bm{Y}^{neg}$ contains $n^{neg}$ terms. Then, the corresponding $\bm{Y}^{pos}$ and $\bm{Y}^{neg}$ are traversed to calculate the similarities $\bm{S^{pos}}$ and $\bm{S^{neg}}$. Specifically, $\bm{S^{pos}}$ is defined by:
\begin{equation}
	\bm{S^{pos}} = \lbrace sim\left({y}_{i}^{pos}, {y}_{k}^{pos}\right) \ | \ {y}_{i}^{pos}, {y}_{j}^{pos} \in \bm{Y}^{pos} \ and \ i \neq j \rbrace
	\label{eq:sim1}
\end{equation}
which can be abbreviated as $\bm{S^{pos}} = \lbrace s_{i}^{pos}\rbrace_{i=1}^{N^{pp}}$, where $sim$ stands for the similarity between features, and $N^{pp}$ denotes $C_{n^{pos}}^2$. $\bm{S^{neg}}$ is defined by:
\begin{equation}
	\bm{S^{neg}} = \lbrace sim\left({y}_{i}^{pos}, {y}_{j}^{neg}\right) \ | \ {y}_{i}^{pos} \in \bm{Y}^{pos} \ and \ {y}_{j}^{neg} \in \bm{Y}^{neg} \rbrace
	\label{eq:sim2}
\end{equation}
which is abbreviated as $\bm{S^{neg}} = \lbrace s_{j}^{neg}\rbrace_{j=1}^{N^{pn}}$, where $N^{pn}$ denotes $n^{pos} \times n^{neg}$.

For the initialized pools, the positive pool cannot include all real positive terms and the negative pool contains certain fake negative terms, since the pools are generated by augmenting the images. In practice, we find a non-negligible issue with the view filter: some fake positives may get transferred to the positive pool during the transfer process of top-k negative terms. To get rid of the fake positives, we further adopt a moving-average mechanism to measure the similarities of samples in the self-adjusted pool (Sap) to avoid such an issue. In detail, for $s_{i}^{pos}$ and $s_{j}^{neg}$, we define:
\begin{equation}
	s_e  =
	\begin{cases}
		\sum_{i = e - \eta}^e \ \ s_i / \eta, & \quad \quad e \geq \eta \\
		\sum_{i = 0}^e \ \ s_i / e, & \quad \quad e < \eta
	\end{cases}
	\label{eq:calcp}
\end{equation}
where $e$ and $i$ denote the epoch numbers, e.g., the $e$-th epoch. Accordingly, $s$ presents $s_{i}^{pos}$ or $s_{j}^{neg}$ of the corresponding epoch. $\eta$ is a hyper-parameter controlling the receptive field of $s$ over epochs. For convenience, $\eta$ is firmly set to 10 on experiments. This mechanism enables IPMC to consider the historical information of $s_{i}^{pos}$ and $s_{j}^{neg}$ so that the view filter can better transfer real positives into the positive pool. The only added burden during the training phase is maintaining a $\eta$-sized memory bank, but for time and space complexity, such burden is slight.

Empirically, we visualize the latent features in Fig. \ref{fig:cltrainplot} by using t-SNE, which proves that IPMC can learn clustered information from multi-view data so that the proposed view filter can transfer real positives to the positive pool. Initially, only feature pairs of different views of the same sample are in the positive pool. As the training of encoders, Sap can transfer fake negatives to the positive pool. \cite{2022Chaos} gives theoretical proof that the contrastive loss can constrain the upper and lower bounds of the cross-entropy loss on downstream tasks, which further proves that the intuition behind our proposed IPMC's behavior is sound, i.e., the fake negatives can be \textit{correctly} selected by the proposed view filter during training. Then, in order to model the multi-view information by contrasting pools, we adopt a novel unified loss.

\begin{table*}[t]
	\tiny
	\renewcommand\arraystretch{1.1}
	\caption{Comparison of different methods on classification accuracy (top 1). We use \textit{conv} and \textit{fc} backbones in the experiments. $^\ddagger$ denotes that the methods have reduced learnable parameters (See Sec. \ref{sec:imageclass}).}
	\label{tab:a}
	\setlength{\tabcolsep}{1.5pt}
	\begin{center}
		\begin{small}
			\begin{tabular}{l|cc||cc||cc||cc}
				\hline
				\multirow{2}*{Model} & \multicolumn{2}{c||}{Tiny ImageNet} & \multicolumn{2}{c||}{STL-10} & \multicolumn{2}{c||}{CIFAR-10} & \multicolumn{2}{c}{CIFAR-100} \\ 
				\cline{2-9}
				& conv & fc & conv & fc& conv & fc& conv & fc \\
				\hline
				\text{Fully supervised} & \multicolumn{2}{c||}{36.60} & \multicolumn{2}{c||}{68.70} & \multicolumn{2}{c||}{75.39} & \multicolumn{2}{c}{42.27} \\
				\hline
				\text{BiGAN \cite{donahue2016adversarial}} & 24.38 & 20.21 & 71.53 & 67.18 & 62.57 & 62.74 & 37.59 & 33.34 \\
				\text{NAT \cite{bojanowski2017unsupervised}} & 13.70 & 11.62 & 64.32 & 61.43 & 56.19 & 51.29 & 29.18 & 24.57 \\
				\text{DIM \cite{hjelm2018learning}} & 33.54 & 36.88 & 72.86 & 70.85 & 73.25 & 73.62 & 48.13 & 45.92 \\
				\text{SplitBrain$^{\ddagger}$ \cite{Splitp2}} & 32.95 & 33.24 & 71.55 & 63.05 & 77.56 & 76.80 & 51.74 & 47.02 \\
				\text{SwAV \cite{2020Mathilde}} & 39.56 $\pm$ 0.2 & 38.87 $\pm$ 0.3 & 70.32 $\pm$ 0.4 & 71.40 $\pm$ 0.3 & 68.32 $\pm$ 0.2 & 65.20 $\pm$ 0.3 & 44.37 $\pm$ 0.3 & 40.85 $\pm$ 0.3 \\
				\text{SimCLR \cite{chen2020simple}} & 36.24 $\pm$ 0.2 & 39.83 $\pm$ 0.1 & 75.57 $\pm$ 0.3 & 77.15 $\pm$ 0.3 & 80.58 $\pm$ 0.2 & 80.07 $\pm$ 0.2 & 50.03 $\pm$ 0.2 & 49.82 $\pm$ 0.3 \\
				\text{CMC$^{\ddagger}$ \cite{Tian2019Contrastive}} & 42.03 $\pm$ 0.2 & 41.09 $\pm$ 0.1 & 83.28 & 86.66 & 81.59 $\pm$ 0.3 & 83.33 $\pm$ 0.2 & 58.71 $\pm$ 0.2 & 57.21 $\pm$ 0.2 \\
				\text{MoCo \cite{2020Kaiming}} & 35.90 $\pm$ 0.2 & 41.37 $\pm$ 0.2 & 77.50 $\pm$ 0.2 & 79.73 $\pm$ 0.3 & 76.37 $\pm$ 0.3 & 79.30 $\pm$ 0.2 & 51.04 $\pm$ 0.2 & 52.31 $\pm$ 0.2 \\
				\text{BYOL \cite{2020Bootstrap}} & 41.59 $\pm$ 0.2 & 41.90 $\pm$ 0.1 & 81.73 $\pm$ 0.3 & 81.57 $\pm$ 0.2 & 77.18 $\pm$ 0.2 & 80.01 $\pm$ 0.2 & 53.64 $\pm$ 0.2 & 53.78 $\pm$ 0.2 \\
				\text{Barlow Twins \cite{2021Barlow}} & 39.81 $\pm$ 0.3 & 40.34 $\pm$ 0.2 & 80.97 $\pm$ 0.3 & 81.43 $\pm$ 0.3 & 76.63 $\pm$ 0.3 & 78.49 $\pm$ 0.2 & 52.80 $\pm$ 0.2 & 52.95 $\pm$ 0.2 \\
				\text{DACL \cite{2021Vikas}} & 40.61 $\pm$ 0.2 & 41.26 $\pm$ 0.1 & 80.34 $\pm$ 0.2 & 80.01 $\pm$ 0.3 & 81.92 $\pm$ 0.2 & 80.87 $\pm$ 0.2 & 52.66 $\pm$ 0.2 & 52.08 $\pm$ 0.3 \\
				\text{LooC \cite{2021Tete}} & 42.04 $\pm$ 0.1 & 41.93 $\pm$ 0.2 & 81.92 $\pm$ 0.2 & 82.60 $\pm$ 0.2 & 83.79 $\pm$ 0.2 & 82.05 $\pm$ 0.2 & 54.25 $\pm$ 0.2 & 54.09 $\pm$ 0.2 \\
				\text{SwAV + Debiased \cite{2020Debiased}} & 39.60 $\pm$ 0.3 & 39.63 $\pm$ 0.3 & 71.29 $\pm$ 0.3 & 72.56 $\pm$ 0.2 & 70.93 $\pm$ 0.3 & 73.81 $\pm$ 0.2 & 51.02 $\pm$ 0.2 & 51.40 $\pm$ 0.2 \\
				\text{SwAV + Hard \cite{2020Hard}} & 41.16 $\pm$ 0.3 & 40.31 $\pm$ 0.3 & 73.55 $\pm$ 0.3 & 74.03 $\pm$ 0.4 & 73.08 $\pm$ 0.3 & 75.67 $\pm$ 0.2 & 51.82 $\pm$ 0.2 & 52.46 $\pm$ 0.2 \\
				\text{SimCLR + Debiased \cite{2020Debiased}} & 38.79 $\pm$ 0.2 & 40.26 $\pm$ 0.2 & 77.09 $\pm$ 0.3 & 78.39 $\pm$ 0.2 & 80.89 $\pm$ 0.2 & 80.93 $\pm$ 0.2 & 51.38 $\pm$ 0.2 & 51.09 $\pm$ 0.2
				\\
				\text{SimCLR + Hard \cite{2020Hard}} & 40.05 $\pm$ 0.3 & 41.23 $\pm$ 0.2 & 79.86 $\pm$ 0.2 & 80.20 $\pm$ 0.2 & 82.13 $\pm$ 0.2 & 82.76 $\pm$ 0.1 & 52.69 $\pm$ 0.2 & 53.13 $\pm$ 0.2 \\
				\text{CMC$^{\ddagger}$ + Debiased \cite{2020Debiased}} & 41.86 $\pm$ 0.2 & 41.61 $\pm$ 0.2 & 83.96 $\pm$ 0.2 & 85.81 $\pm$ 0.2 & 82.29 $\pm$ 0.2 & 83.75 $\pm$ 0.2 & 59.04 $\pm$ 0.2 & 57.66 $\pm$ 0.2 \\
				\text{CMC$^{\ddagger}$ + Hard \cite{2020Hard}} & 42.93 $\pm$ 0.2 & 42.56 $\pm$ 0.3 & 83.81 $\pm$ 0.3 & \textbf{87.15 $\pm$ 0.2} & 83.44 $\pm$ 0.2 & 86.31 $\pm$ 0.3 & 59.32 $\pm$ 0.2 & 59.33 $\pm$ 0.2 \\
                    {\text{SimSiam \cite{2021simsiam}}} & {41.03 $\pm$ 0.3} & {41.27 $\pm$ 0.3} & {80.91 $\pm$ 0.2} & {81.88 $\pm$ 0.2} & {78.14 $\pm$ 0.3} & {81.13 $\pm$ 0.2} & {52.55 $\pm$ 0.2} & {53.52 $\pm$ 0.2} \\
                    {\text{CoCoNet \cite{li2022modeling}}} & {42.28 $\pm$ 0.2} & {\textbf{43.63 $\pm$ 0.2}} & {\textbf{85.34 $\pm$ 0.1}} & {83.82 $\pm$ 0.2} & {83.10 $\pm$ 0.3} & {83.24 $\pm$ 0.2} & {58.64 $\pm$ 0.2} & {58.21 $\pm$ 0.3} \\
                    {VICReg \cite{DBLP:conf/nips/BardesPL22}} & {41.08 $\pm$ 0.2} & {41.89 $\pm$ 0.3} & {80.61 $\pm$ 0.3} & {80.93 $\pm$ 0.3} & {79.51 $\pm$ 0.3} & {81.84 $\pm$ 0.3} & {53.95 $\pm$ 0.3} & {53.05 $\pm$ 0.3} \\
				\hline
				\textbf{IPMC(\textit{Fp})$^\ddagger$} & 43.91 $\pm$ 0.2 & 41.51 $\pm$ 0.2 & 83.70 $\pm$ 0.2 & 86.81 $\pm$ 0.2  & 84.84 $\pm$ 0.2 & 85.99 $\pm$ 0.3 & 59.05 $\pm$ 0.2 & 58.95 $\pm$ 0.2 \\
				\textbf{IPMC(\textit{Fp} + \textit{DA})$^\ddagger$} & 45.01 $\pm$ 0.2 & 41.55 $\pm$ 0.2 & 83.90 $\pm$ 0.3 & 86.92 $\pm$ 0.2 & 84.86 $\pm$ 0.2 & 87.97 $\pm$ 0.2 & 59.89 $\pm$ 0.2 & 60.07 $\pm$ 0.2 \\
				\textbf{IPMC(\textit{Sap} + \textit{DA})$^\ddagger$} & \textbf{45.11 $\pm$ 0.2} & 42.99 $\pm$ 0.2 & 84.11 $\pm$ 0.2 & 86.94 $\pm$ 0.2 & \textbf{84.90 $\pm$ 0.3} & \textbf{88.29 $\pm$ 0.2} & \textbf{60.12 $\pm$ 0.2} & \textbf{60.58 $\pm$ 0.2} \\
				\hline
			\end{tabular}
		\end{small}
	\end{center}
\end{table*}

\subsection{Instance-tier: a unified loss}
As demonstrated in Fig. \ref{fig:algoframe}, motivated by \cite{FaceNet2}, we formulate the unified loss of our model using the acquired $\bm{S^{pos}}$ and $\bm{S^{neg}}$ as follows:
\begin{align}
	\mathcal{L} = \left[\bm{S^{neg}}-\bm{S^{pos}}+\lambda \right]_{+} ,
	\label{eq:L}
\end{align}%
{where $[\cdot]_{+}$ denotes the \textit{cut-off at zero} operation to ensure $\mathcal{L} \geq 0$. $\lambda$ is a margin to enhance the separation between $\bm{S^{pos}}$ and $\bm{S^{neg}}$. While the difference between $\bm{S^{neg}}$ and $\bm{S^{pos}}$ is not the larger the better, the margin $\lambda$ leads to preferable convergence. Further increasing the difference may undermine the final convergence in optimization. Therefore, we adopt the temperature coefficient and the softmax function into the above formula, which can guide to the desirable convergence and reduce the computational intensity in optimization. The reformulated loss function is defined by:}
\begin{equation}
	\begin{aligned}
	\mathcal{L}=\frac{1}{\gamma}log\Bigg\{1+\sum_{i=1}^{N^{pp}}\sum_{j=1}^{N^{pn}}exp\bigg[\gamma(s_{j}^{neg}-s_{i}^{pos}+\lambda)\bigg]\Bigg\}.
	\label{eq:LF}
	\end{aligned}
\end{equation}%
{When $\gamma \to +\infty$, Eq. \ref{eq:LF} is exactly approximated by Eq. \ref{eq:L}. Inspired by \cite{Circle2}, we add the leveraging factors $\alpha^{pos}$ and $\alpha^{neg}$ to modulate the weights of $s^{pos}$ and $s^{neg}$. $\alpha^{pos}$ and $\alpha^{neg}$ jointly amplify the impact of the instance similarity that deviates far from the optimal and weaken the impact of the instance similarity that is close to the optimal. Thus, the loss can lay emphasis on optimizing the instance similarity (i.e., $s^{pos}$ or $s^{neg}$) that can make a greater contribution to optimization. We add interval factors $\delta^{pos}$ and $\delta^{neg}$ in Eq.\ref{eq:LF} to substitute $\lambda$:}
\begin{equation}
	\begin{aligned}
	&\mathcal{L}_{UniSap} = \frac{1}{\gamma}log\Bigg\{1+\sum_{i=1}^{N^{pp}}\sum_{j=1}^{N^{pn}}exp\bigg[\gamma\Big(\\&\alpha^{neg}(s_{i}^{neg}-\delta^{neg})- \alpha^{pos}(s_{j}^{pos}-\delta^{pos})\Big)\bigg]\Bigg\},
	\label{eq:LFa2}
	\end{aligned}
\end{equation}%
{where $\alpha^{pos} = [O^{pos}-s^{pos}_i]_{+}$ and $\alpha^{neg} = [s^{neg}_j-O^{neg}]_{+}$, where $O^{neg}$ and $O^{pos}$ represents the optimums of $s^{neg}_j$ and $s^{pos}_i$, respectively. $\delta^{neg}$ may equal to $\delta^{pos}$. Inspired by \cite{Circle2}, by normalizing the features of $\bm{Y}^{neg}$ and $\bm{Y}^{pos}$, we limit the values of $s^{neg}$ and $s^{pos}$ to $[0, 1]$. To optimize $s^{neg}$ to 0 and $s^{pos}$ to 1 and cut the number of hyper-parameters, we set $O^{pos}=1+\delta$, $O^{neg}=-\delta$, $\delta^{pos}=1-\delta$, and $\delta^{neg}=\delta$. Integrate interval factors into Eq. \ref{eq:LFa2}, we obtain:}
\begin{equation}
	\begin{aligned}
	&\mathcal{L}_{UniSap} = \frac{1}{\gamma}log\Bigg\{1+\sum_{i=1}^{N^{pp}}\sum_{j=1}^{N^{pn}}\\&exp\bigg[\gamma\Big({(s_{j}^{pos}-1)}^2+{(s_{i}^{neg})}^{2}-2\delta^{2}\Big)\bigg]\Bigg\},
	\label{eq:Luni}
	\end{aligned}
\end{equation}%
{where the decision boundary of similarities is depicted as ${(s^{pos}-1)}^2+{(s^{neg})}^{2}=2 \times \delta^{2}$, and only two hyper-parameters, i.e., $\gamma$ and $\delta$, are preserved. The mechanism behind the behavior in Eq. \ref{eq:Luni} can be treated as the process to promote $s^{pos}$ approaching 1 and $s^{neg}$ approaching 0 with the decision boundary restricted by the radius $\delta$. Considering $\delta$, such a loss function actually aims to achieve that $s^{pos} > 1 - \delta$ while $s^{neg} < \delta$, and when $\delta$ is approaching 0, the aforementioned purpose can be acquired.}

In practice, we find that directly applying $\mathcal{L}_{UniSap} $ in our method causes the loss to converge excessively fast into a local minimum due to the introduction of the leveraging factor $\alpha^{pos}$ and $\alpha^{neg}$. Therefore, we propose the unified loss by adapting $\alpha$ to $\bar{\alpha} = [\alpha^{\tau_{dec}} / {\phi_{dec}} + 1]_{+}$, where $\phi_{dec}$ is a linear attenuation coefficient to attenuate the impact of $\alpha$ so that the difference between the current value and the optimum becomes smaller, and $\tau_{dec}$ is an exponential coefficient to nonlinearly adjust the impact of $\alpha$. Our loss is more sensitive to similarities that are far from the optimum when $\tau_{dec}$ becomes larger, i.e., $\alpha$'s impact is amplified by $\tau_{dec}$. We further conduct parameter experiments to derive the appropriate $\phi_{dec}$ and $\tau_{dec}$, which is demonstrated in Sec. \ref{sec:imageclass}.

By jointly using the self-adjusted pool contrast and the unified loss, IPMC can maintain the sufficiency of self-supervision, i.e., $I(Y^*;V_1;V_2) = I(X;V_1;V_2)$.

\begin{table}[t]
    \vskip 0.34in
    \caption{Comparison of image classification accuracy (top 1) on ImageNet.}
    \label{tab:b}
    \setlength{\tabcolsep}{5.5pt}
    \begin{center}
        \begin{small}
            \begin{tabular}{l|c}
                \hline
                \multicolumn{2}{c}{ImageNet} \\ 
                \cline{1-2} Method & conv \\
                \hline
                \text{Fully supervised} & 50.5 \\
                \hline
                \text{DeepCluster \cite{2018Caron}} & 36.1 \\
                \text{SwAV \cite{2020Mathilde}} & 38.0 $\pm$ 0.3 \\
                \text{SimCLR \cite{chen2020simple}} & 37.7 $\pm$ 0.2 \\
                \text{CMC$^\ddagger$ \cite{Tian2019Contrastive}} & 42.8 \\
                \text{MoCo \cite{2020Kaiming}} & 39.4 $\pm$ 0.2 \\
                \text{BYOL \cite{2020Bootstrap}} & 41.1 $\pm$ 0.2 \\
                \text{Barlow Twins \cite{2021Barlow}} & 39.6 $\pm$ 0.2 \\
                \text{DACL \cite{2021Vikas}} & 41.8 $\pm$ 0.2 \\
                \text{LooC \cite{2021Tete}} & 43.2 $\pm$ 0.2 \\
                \text{SwAV + Debiased \cite{2020Debiased}} & 39.3 $\pm$ 0.3 \\
                \text{SwAV + Hard \cite{2020Hard}} & 42.9 $\pm$ 0.3 \\
                \text{SimCLR + Debiased \cite{2020Debiased}} & 38.9 $\pm$ 0.3 \\
                \text{SimCLR + Hard \cite{2020Hard}} & 41.5 $\pm$ 0.2 \\
                \text{CMC$^\ddagger$ + Debiased \cite{2020Debiased}} & 42.9 $\pm$ 0.2 \\
                \text{CMC$^\ddagger$ + Hard \cite{2020Hard}} & 43.3 $\pm$ 0.3 \\
                {\text{SimSiam \cite{2021simsiam}}} & {41.9 $\pm$ 0.3} \\
                {\text{CoCoNet \cite{li2022modeling}}} & {43.8 $\pm$ 0.1} \\
                {\text{VICReg \cite{DBLP:conf/nips/BardesPL22}}} & {42.5 $\pm$ 0.3} \\
                \hline
                \textbf{IPMC(\textit{Fp})$^\ddagger$} & 43.8 $\pm$ 0.2 \\
                \textbf{IPMC(\textit{Fp} + \textit{DA})$^\ddagger$} & 44.1 $\pm$ 0.2 \\
                \textbf{IPMC(\textit{Sap} + \textit{DA})$^\ddagger$} & \textbf{44.6 $\pm$ 0.3} \\
                \hline
            \end{tabular}
        \end{small}
    \end{center}
\end{table}

\subsection{Model objective}
We incorporate the objectives of view distribution alignment and self-adjusted contrastive learning  into:
\begin{equation}
	\label{equal16}
	\mathcal{L}_{IPMC} = {\beta \cdot \mathcal{L}_{DA}} + {\mathcal{L}_{UniSap}}
\end{equation}
where $\beta$ is the coefficient that controls the balance between $\mathcal{L}_{DA}$ and $\mathcal{L}_{UniSap}$. The overall objective of IPMC only has three hyper-parameters, i.e., $\beta$, $\gamma$, and $\delta$. We conduct experiments to study their impact in Sec. \ref{sec:deepexplore}.

\begin{figure}[t]
	\begin{center}
		\includegraphics[width=0.4\textwidth]{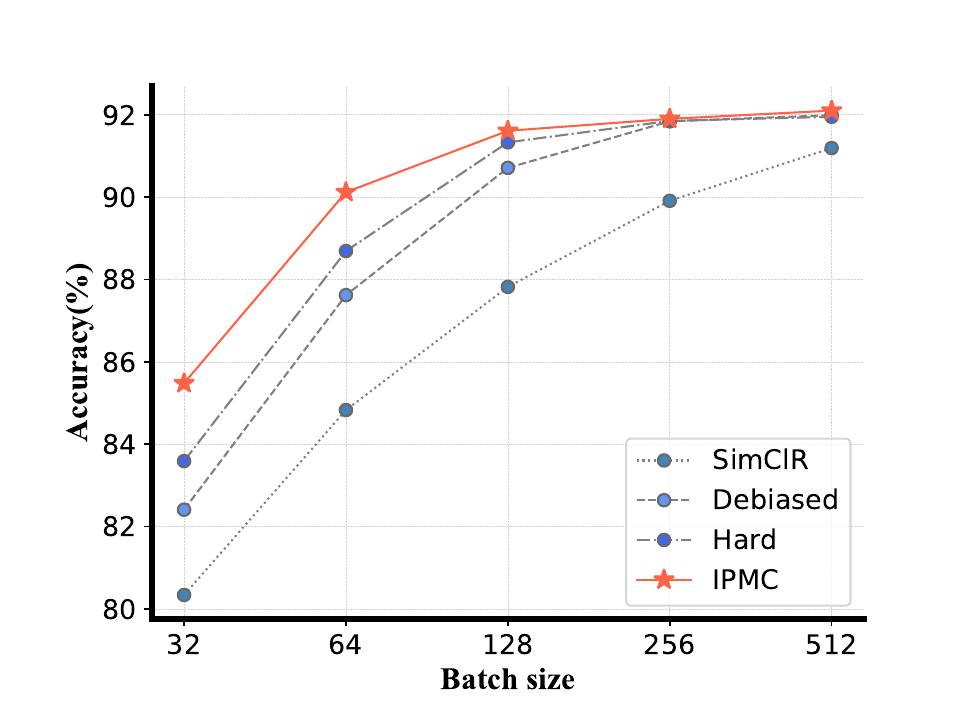}
		\caption{Comparison of image classification accuracy (top 1) on CIFAR-10 with ResNet50, which was conducted by following the settings of Hard \cite{2020Hard}.}
		\label{fig:batchplot}
		\vspace{0.5cm}
	\end{center}
\end{figure}

\begin{figure*}
    \begin{center}
        \centerline{\includegraphics[width=1.95\columnwidth]{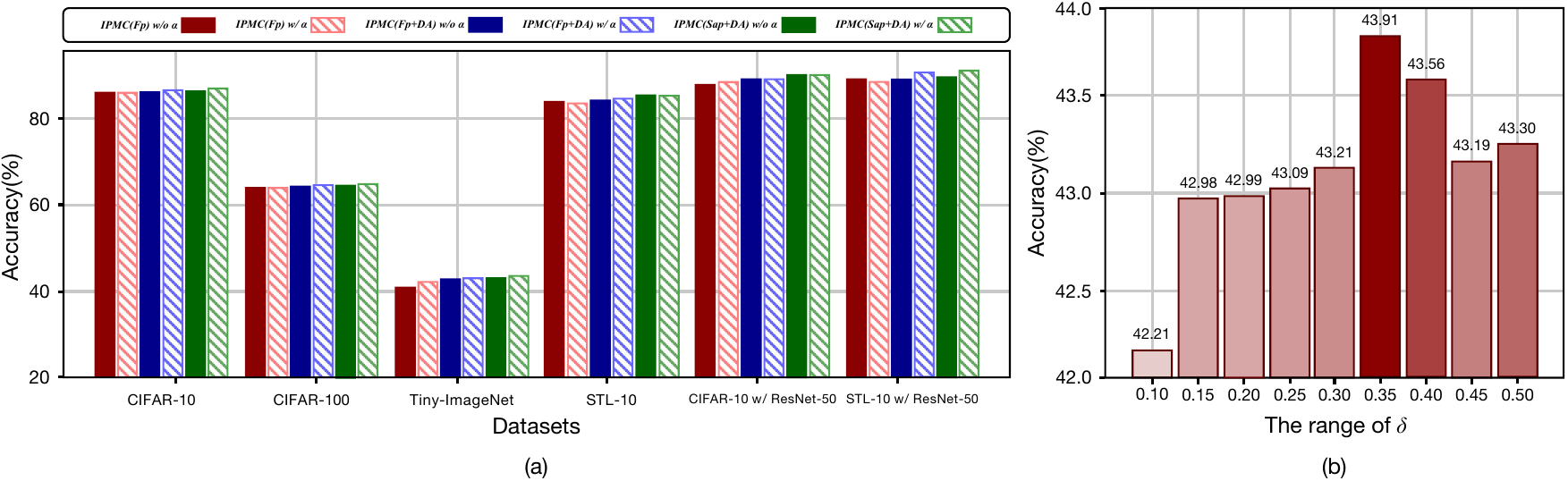}}
        \vskip -0.13in
        \caption{Exploration of the leveraging factor $\alpha$ and interval factor $\delta$ with \textit{conv} encoders. (a) manifests the evaluations of our models with $\alpha$ or without $\alpha$. We further employed our methods with ResNet-50 \cite{2016Kaiming} on CIFAR-10 and STL-10. (b) shows the effect of different $\delta$, and the comparisons are conducted on CIFAR-10 benchmark dataset by using the ablation variant IPMC(\textit{Fp}).}
        \label{fig:alphadeltastats}
    \end{center}
\end{figure*}

\section{Experiments}\label{sec:experiments}
To effectively evaluate the performance and transferability of IPMC, we conducted several comparisons on benchmark datasets. The deepgoing exploration is further conducted to clarify the property of our method.

\subsection{Image classification comparisons} \label{sec:imageclass}
\subsubsection{Preparation}
We benchmarked our IPMC on five established datasets, i.e., Tiny ImageNet \cite{krizhevsky2009learning}, STL-10 \cite{coates2011analysis}, CIFAR-10 \cite{krizhevsky2009learning}, CIFAR-100 \cite{krizhevsky2009learning} and ImageNet \cite{2009Feifei}, within three backbone networks. Specifically, in Tab. \ref{tab:a} and \ref{tab:b}, \textit{conv} depicts that the encoder with the 5 convolutional layers in Alexnet is adopted as the backbone, and \textit{fc} represents the utilization of the encoder with the 5 convolutional layers and 2 fully connected layers in Alexnet. In the experiments of Tab. \ref{fig:batchplot} and several experiments of Tab. \ref{fig:alphadeltastats} (a), we use ResNet-50 \cite{2016Kaiming} as the encoders. We compared IPMC against a fully-supervised method (similar to Alexnet \cite{2012Krizhevsky}) and the state-of-the-art unsupervised methods. We also performed the ablation studies by removing the distribution-tier and the dynamic adaptation of the contrasting pool. Specifically, \textit{Fp} denotes the vanilla \textit{fixed} pool, and \textit{Sap} denotes the proposed \textit{self-adjusted} pool. \textit{DA} denotes the view distribution alignment. We followed the basic experimental settings (e.g., batch size, etc.) of CMC \cite{Tian2019Contrastive}. For effective verification, we selected three views: the Red-Green-Blue (RGB) view of the original image, the luminance channel (L) view, and the ab-color channel (ab) view in the comparisons of Tab. \ref{tab:a}. In the comparisons of Tab. \ref{tab:b}, we adopted Chroma Subsampling (i.e., YDbDr) views of an image for the multi-view setting. For a fair comparison, we adopted the same data augmentation methods as CMC \cite{Tian2019Contrastive} (i.e., random crop and horizontal flip). According to the view pool setting, we took 4096 images as the negative pools for each positive pool. Meanwhile, a conventional memory bank \cite{un2} is adopted to facilitate calculations with storing learned features. We therefore can efficiently retrieve the other 4096 negative pools from the memory bank to pair with the corresponding positive pools, and it is not needed to recompute the corresponding features. We instantaneously updated the memory bank when computing the features. Then we evaluated the performance of models by averaging the results of the last 100 epochs of optimizations. Also, to alleviate the over-fitting problem on the test set across models, we uniformly set the learning rates, dropout rates, and weight decay rates. In the experiment, the built deep learning representation from multiple views provides outstanding performance, which outperforms the state-of-the-art methods.

We collected the results of 20 trials for comparisons. The average result of the last 20 epochs is used as the final result of each trial. The average results from total of 20 trials are presented in tables, and the 95\% confidence intervals are also reported. The results without 95\% confidence intervals are quoted from the original papers.
			
\subsubsection{Classification results and discussion}
Tab. \ref{tab:a} and \ref{tab:b} show the comparisons on five benchmark datasets. The last three rows of tables represent the results of our proposed methods. On average, IPMC(\textit{Sap} + \textit{DA}) beats the best prior methods on all datasets. Generally, CMC outperforms many remarkable state-of-the-art methods, which may due to that the architecture of CMC can better explore the shared information among multiple views (especially more than one). To the best of our knowledge, in the field of unsupervised learning, the results of IPMC are state-of-the-art. The IPMC results indicate a relatively large performance improvement when compared with the fully-supervised method trained end-to-end (without fine-tuning) for the architecture presented, which demonstrates that the representations learned by IPMC are better. 

From the perspective of data augmentation, we reckon the reason CMC can outperform most benchmark methods with the mentioned settings is that several methods, e.g., SimCLR and BYOL, use the same weak augmentation to generate views since they predict a view by another so the difference between views should be small and the large batch size is required, while CMC uses different channels of color spaces as views (can be treated as strong data augmentations), thus the informativeness of such views is much larger so that CMC can outperform others with small batch sizes. We follow the setting of CMC so that the performance of SimCLR and BYOL may degenerate, because the adopted views are generated by different data augmentations, which is contrary to the requirement of specific methods, e.g., SimCLR, BYOL, etc. However, for the multi-view methods, e.g., CMC and IPMC, when the more powerful backbone networks are used as encoders and sophisticated data augmentation methods are adopted, the unsupervised learning approaches have increasingly outstanding performance.

\begin{figure}
	\centering
	\includegraphics[width=0.48\textwidth]{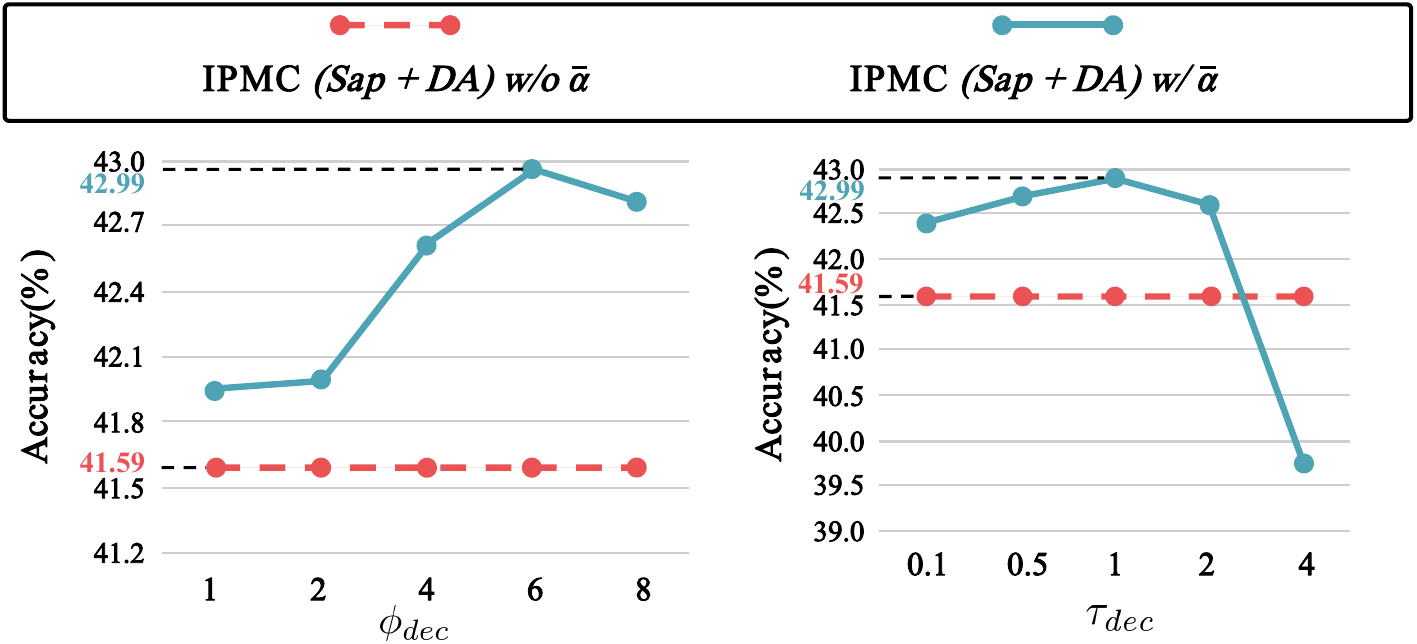}
	\caption{Comparison of image classification accuracy (top 1) on Tiny ImageNet with \textit{fc} encoder to evaluate the impact of $\phi_{dec}$ and $\tau_{dec}$ on IPMC.}
	\label{fig:leveragingfactor}
	\vspace{0.5cm}
\end{figure}

\subsubsection{Study on ablation models} \label{sec:ablresults}
As demonstrated in Tab. \ref{tab:b}, the ablation models outperform most of the state-of-the-art approaches but fall short when compared to IPMC(\textit{Sap} + \textit{DA}). On average, IPMC(\textit{Sap} + \textit{DA}) and IPMC(\textit{Fp} + \textit{DA}) outperform IPMC(\textit{Fp}), which supports that each of the proposed techniques has a positive impact on IPMC's performance.

\begin{figure*}
    \begin{center}
        \centerline{\includegraphics[width=1.95\columnwidth]{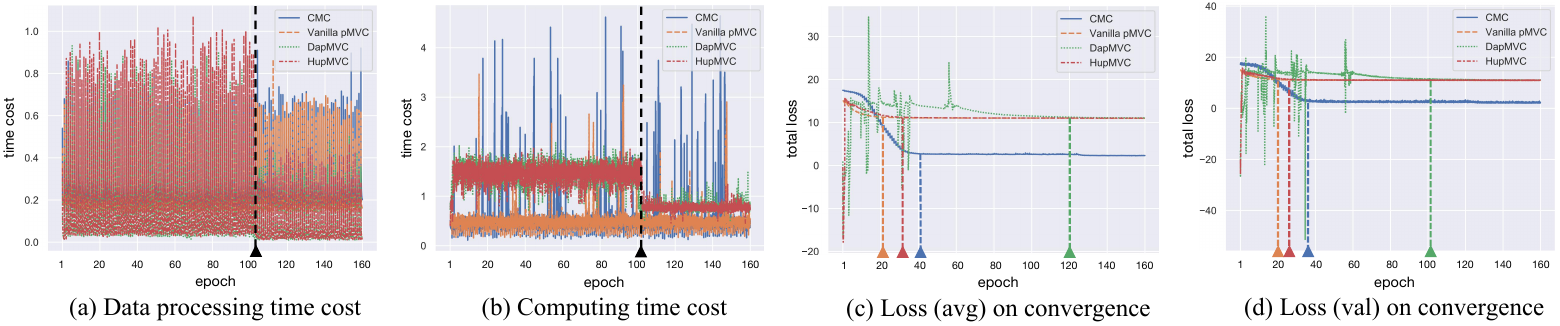}}
        \caption{Optimization analyses of our proposed methods and CMC on CIFAR-100. (a) demonstrates the time cost of data processing, including data loading, data augmentations, etc. (b) demonstrates the computational time cost of the feed-forward calculation and back-propagation training of the encoders. (c) and (d) show the moving average and instant value of the total loss, respectively.}
        \label{fig:convergence}
    \end{center}
\end{figure*}

\subsubsection{Performing IPMC with different batch sizes}
As demonstrated in Fig. \ref{fig:batchplot}, we compared our proposed IPMC with three benchmark methods with different batch sizes by following the experimental settings of Hard \cite{2020Hard}. The reported results prove that IPMC outperforms the compared methods with different batch sizes by using ResNet-50. As the batch size increases, the improvement of our method to benchmark methods gradually becomes smaller, but even in the case of the batch size being 512, our method can still outperform other methods. 

\subsubsection{Study on leveraging factor $\alpha$}
Fig. \ref{fig:alphadeltastats} (a) shows the details of the experiments on leveraging factor $\alpha$. The result implies that $\alpha$ indeed leverages the biases of the similarities. The IPMC models with $\alpha$ averagely outperform the models without $\alpha$ by 0.74\% on Tiny ImageNet, 0.78\% on CIFAR-10, and 0.21\% on CIFAR-100, respectively. These results further support the advance of taking $\alpha$. Yet the models without $\alpha$ beat the $\alpha$-based models by 0.23\% on STL-10. We conjecture that the labeled data of STL-10 is relatively small, which may make the experimental results unstable, since it only has 5,000 labeled images while CIFAR-10 has 45,000 labeled images. Therefore, the derived results of STL-10 might be relatively inconsistent. The results on STL-10 w/ ResNet-50 prove the inconsistency of the experiments conducted on STL-10 dataset.

As shown in Fig. \ref{fig:leveragingfactor}, $\bar{\alpha}$ can improve IPMC. Specifically, when $\phi_{dec} = 6$ and $\tau_{dec} = 1$, our IPMC can achieve the best performance, which indicates that compared with adopting $\alpha$ as $\alpha^{neg}$ and $\alpha^{pos}$, $\bar{\alpha}$ can further improve IPMC by using appropriate settings of $\phi_{dec}$ and $\tau_{dec}$. Comparing the \textcolor[RGB]{100,150,235}{blue} curves with the \textcolor[RGB]{255,100,30}{red} curves, we observe that IPMC with adopting $\bar{\alpha}$ has better performance, which proves the effectiveness of $\bar{\alpha}$.

\subsubsection{Study on interval factor $\delta$}
Fig. \ref{fig:alphadeltastats} (b) shows the evaluation results on the influence of the interval factor $\delta$. We observe that an appropriate chosen $\delta$ (e.g., 0.35) can improve IPMC by at least 1.70\% (compared with the result derived when $\delta$ is equal to 0.10), which supports our conjecture that $\delta$ helps to enhance the discriminability of positive and negative pools by inserting an interval between the similarities.

\subsubsection{Optimization analyses} \label{sec:opt}
As shown in Fig. \ref{fig:convergence} (a) and (b), IPMC(\textit{Fp}) has consistent and lower costs in optimization, because our method jointly uses the pool architecture and memory bank to alleviate computational intensity in the set-tier. The view distribution alignment causes an increase in costs. Yet, when the distributions are already aligned, the costs bounce back to the normal level. The distribution alignment assists encoders to efficiently reduce the view-specific noise, which is revealed by the decrease of data processing time cost after around 100 epochs in Fig. \ref{fig:convergence} (a). Fig. \ref{fig:convergence} (c) and (d) indicate that IPMC(\textit{Fp}) can accelerate the convergence due to the unified loss calculation in the instance-tier. Additionally, the self-adjusted pool helps to tackle the optimization fluctuation.

\subsubsection{Discussion on the simplicity of our method}
For the simplicity of the encoder, we follow the network splitting of \cite{un2} so that our model is significantly smaller than most benchmark models. The reason behind the simplicity of IPMC is related to the adoption of network splitting. According to the principle of building the encoders, the AlexNet is split across the channel dimension with a conjecture that split-AlexNet can also perform well in learning representations between views, and the split-AlexNet only has the halved learnable parameters \cite{Splitp2}. We, therefore, built the AlexNet with 5 convolutional layers (attached with auxiliary batchnorm layers, ReLU activation functions, and corresponding maxpool functions), 2 linear layers (with corresponding batchnorm layers and ReLU activation functions), and a fully connected layer followed by a l2 normalization function, which is to tackle the problem of distribution drift, and then the split-AlexNets (i.e., the sub-networks) are served as the encoders. In experiments, we used the conv network and fc network, which use the corresponding layers of AlexNet (note that we split across channels for RGB, L, and ab views), as the encoders. In training, we hold the perspective that the representations learned the crucial features of views through different encoders. In the test, we directly concatenated representations layer-wise from the encoders into one in order to achieve the ultimate representation of an input sample.

For the simplicity of the classifier, we directly leverage a basic linear network followed by a softmax output function as the classifier on downstream tasks. Following the proposed experimental setting of the previous literature \cite{2018RepresentationOord, hjelm2018learning, 2019Arora, Tian2019Contrastive}, we evaluated the quality of the learned representations by freezing the weights of backbone encoders and training linear classifiers (adopted on all tasks) on top of each layer.

For building the discrepancy metric calculation critic network based on Wasserstein distance (i.e., the critic network), the discrepancy metric of IPMC is to measure the differences between views in the learned latent space. We also consider the simplicity of the critic network, which measures the differences and is designed with four linear layers followed by three ReLU activation functions, and the first hidden layer consists of 1,000 units. The implementations of the Lipschitz criteria work in the same way as \cite{2017Shen}.

\begin{table}[t]
	\renewcommand\arraystretch{1.1}
	\vskip 0.5in
	\caption{Action recognition accuracy (\%) to evaluate \textit{task} and \textit{dataset} transferability on benchmark video datasets. We followed the setting of \cite{Tian2019Contrastive, 2007Christopher}. $\ast$ denotes our reimplementation.}
	\label{tab:action}
	\setlength{\tabcolsep}{2.1pt}
	\begin{center}
		\begin{small}
			\begin{tabular}{l|c|c|c}
				\hline
				\text{Method} & \text{Views} & \text{UCF-101} & \text{HMDB-51} \\
				\hline
				\text{Random} & - & 48.2 & 19.5 \\
				\text{ImageNet} & - & 67.7 & 28.0 \\
				\hline
				\text{TempCoh \cite{DBLP:conf/icml/MobahiCW09}} & 1 & 45.4 & 15.9 \\
				\text{Shuffle and Learn \cite{DBLP:conf/eccv/MisraZH16}} & 1 & 50.2 & 18.1 \\
				\text{Geometry \cite{2018Geometry}} & 2 & 55.1 & 23.3 \\
				\text{OPN \cite{2017Unsupervised}} & 1 & 56.3 & 22.1 \\
				\text{ST Order \cite{Uta2018Improving}} & 1 & 58.6 & 25.0 \\
				\text{Cross and Learn \cite{2018Cross}} & 2 & 58.7 & \textbf{27.2} \\
				\text{CMC (only V) \cite{Tian2019Contrastive}} & 2 & 55.3 & - \\
				\text{CMC (only D) \cite{Tian2019Contrastive}} & 2 & 57.1 & - \\
				\text{CMC (V + D) \cite{Tian2019Contrastive}} & 3 & 59.1 & 26.7 \\
				\text{CMC$^\ast$ (V + D) \cite{Tian2019Contrastive}} & 3 & 58.8 & 26.3 \\
				\hline
				\textbf{IPMC} (only V) & 2 & 56.2 & - \\
				\textbf{IPMC} (only D) & 2 & 58.5 & - \\
				\textbf{IPMC} (V + D) & 3 & \textbf{59.5} & 26.7 \\
				\hline
			\end{tabular}
		\end{small}
	\end{center}
\end{table}

\subsection{Action recognition comparisons} \label{sec:action}
\subsubsection{Preparation}
We conducted comparisons on the task of action recognition by following the experimental setting of \cite{Tian2019Contrastive, 2007Christopher}, which is based on video data. To evaluate the performance of our method, we performed IPMC based on the architecture of CMC \cite{Tian2019Contrastive}. We trained our methods on UCF-101 \cite{2012UCF101} by using CaffeNets \cite{2012Krizhevsky} to learn features from video data. Two streams are applied in the method: 1) the ventral (V) stream, which contains a view of a neighbouring frame of the target frame (image) in the video; 2) the dorsal (D) stream, which contains the optical flow (centered at the target frame) in video data as a view.

In the training, we adopted both ventral and dorsal streams, which can be treated as two views, and the target frame in a video stream is the third view. In the test, the compared methods are tested on UCF-101 to evaluate the \textit{task} transferability and on HMDB-51 \cite{2011HMDB} to evaluate the \textit{task} and \textit{dataset} transferability. We performed our method based on the reimplemented CMC, i.e., CMC$^\ast$, and the compared method IPMC is the complete variant, i.e., IPMC(\textit{Sap + DA}).

\begin{table}[t]
	\small
	\renewcommand\arraystretch{1.1}
	\vskip 0.5in
	\caption{Performance (accuracy) on the CIFAR-10 and CIFAR-100 datasets with \textit{fc} encoder. We illustrate the impact of different discrepancy metrics on our proposed method.}
	\label{tab:e}
	\setlength{\tabcolsep}{5.6pt}
	\begin{center}
		\begin{tabular}{l|ccc}
			\hline
			\text{Model} & CIFAR-10 & CIFAR-100 & Average\\
			\hline
			\text{IPMC(\textit{Fp})} & 85.99 & 58.95 & 72.47 \\
			\hline
			\textbf{IPMC(\textit{Fp} + \textit{DA}) - KL} & 86.57 & 59.76 & 73.12 \\
			\textbf{IPMC(\textit{Fp} + \textit{DA}) - WD} & \textbf{88.29} & \textbf{60.58} & \textbf{74.44} \\
			\hline
		\end{tabular}
	\end{center}
\end{table}

\begin{figure*}
	\centering
	\includegraphics[width=1.8\columnwidth]{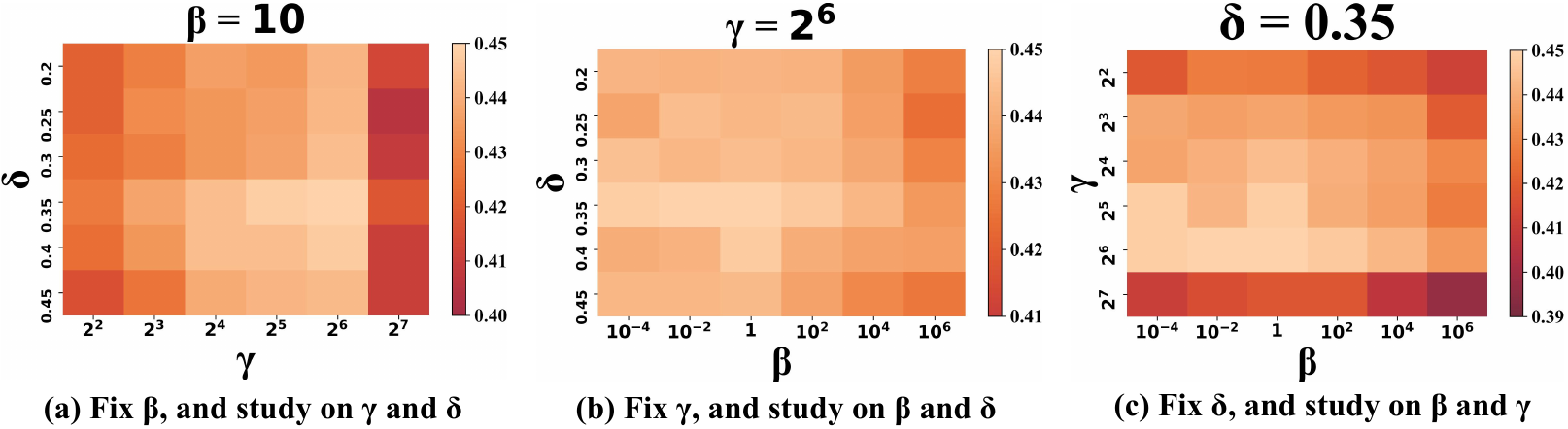}
	\caption{Impacts of the hyper-parameters $\beta$, $\gamma$, and $\delta$ of our proposed method. We conducted comparisons based on IPMC(\textit{Sap} + \textit{DA}) on Tiny ImageNet with \textit{conv} encoder (as in Tab. \ref{tab:a}). In order to measure the influences, we iteratively fixed one parameter and then study on the others by selecting them in the ranges, respectively.}
	\label{fig:heatmap}
	\vspace{0.5cm}
\end{figure*}

\subsubsection{Action recognition results and discussion}
As shown in Tab. \ref{tab:action}, IPMC achieves the state-of-the-art on the action recognition task of video data. Comparing the results on UCF-101, we observe that IPMC has remarkable \textit{task} transferability, since the tasks are different in the training and test phases, and IPMC outperforms the benchmark methods. Comparing the results on HMDB-51, we find that our method has the relatively good \textit{task} and \textit{dataset} transfer-abilities.

However, our method falls short when it is compared with Cross and Learn. We reckon that the views adopted by Cross and Learn are different from that of CMC, and our proposed IPMC is implemented based on CMC$^\ast$ (the reimplementation of CMC). We further compare CMC and our method and observe that IPMC can improve CMC in various settings of the adopted multi-view, i.e., V and D. Therefore, on the action recognition task of video data, our proposed IPMC can still effectively model multi-view data.

\begin{table}[t]
	\small
	\renewcommand\arraystretch{1.1}
	\vskip 0.3in
	\caption{Performance (accuracy) on the Tiny ImageNet and STL-10 datasets with \textit{conv} encoder. To illustrate our theory of multi-view learning. We conducted several experiments based on the \textit{conv} encoder and classifier as in Tab. \ref{tab:a}. The views including the optical RGB view (RGB), the luminance channel view (L), and the ab-color channels view (ab), and we separately grouped the views to introduce them in IPMC(\textit{Sap} + \textit{DA}). Notably, the RGB-L-ab views-based IPMC outperforms other comparison methods.}
	\label{tab:f}
	\setlength{\tabcolsep}{6pt}
	\begin{center}
		\begin{tabular}{l|c|c|c}
			\hline
			\text{Model} & Tiny ImageNet & STL-10 & Average\\
			\hline
			\textbf{IPMC w/ RGB-L-ab} & \textbf{45.11} & \textbf{84.11} & \textbf{64.61} \\
			\hline
			\text{IPMC w/ RGB-L} & 43.69 & 83.61 & 63.65 \\
			\text{IPMC w/ RGB-ab} & 42.38 & 82.85 & 62.62 \\
			\text{IPMC w/ L-ab} & 43.01 & 83.03 & 63.02 \\
			\text{IPMC w/ RGB} & 41.47 & 82.76 & 62.12 \\
			\text{IPMC w/ L} & 37.98 & 75.39 & 56.69 \\
			\text{IPMC w/ ab} & 38.21 & 77.08 & 57.65 \\
			\hline
			\text{CMC w/ RGB-L-ab} & 42.03 & 83.28 & 62.66 \\
			\hline
		\end{tabular}
	\end{center}
\end{table}

\begin{table}[t]
        \small
	\renewcommand\arraystretch{1.1}
	\vskip 0.3in
	\caption{{Data perturbation robustness comparisons of benchmark SSL methods and the proposed IPMC on the Tiny ImageNet dataset, which is performed by implementing different data perturbations on candidate methods. Note that the comparisons are based on the fc backbone.}}
	\vskip -0.in
	\label{tab:diffaug}
	\setlength{\tabcolsep}{5.8pt}
	\begin{center}
		\begin{tabular}{ccccc|cc}
			\hline
			\multicolumn{5}{c|}{Data perturbations} & \multicolumn{2}{c}{Methods} \\
			\hline
		    \multirow{2}*{rotate} & random & random & color & random & \multirow{2}*{CMC} & \multirow{2}*{IPMC} \\
			& crop & grey & jitter & mask & & \\
			\hline
			$\checkmark$ & & & & & 34.51 & \textbf{35.08} \\
			& & & $\checkmark$ & & 36.72 & \textbf{38.14} \\
			& & & & $\checkmark$ & 35.85 & \textbf{36.79} \\
			& $\checkmark$ & & & $\checkmark$ & 35.92 & \textbf{37.03} \\
			& $\checkmark$ & $\checkmark$ & & & 36.69 & \textbf{37.24} \\
			$\checkmark$ & $\checkmark$ & & $\checkmark$ & & 39.21 & \textbf{39.90} \\
			$\checkmark$ & $\checkmark$ & $\checkmark$ & $\checkmark$ & $\checkmark$ & 40.88 & \textbf{43.12} \\
			$\checkmark$ & $\checkmark$ & $\checkmark$ & $\checkmark$ & & 41.09 & \textbf{42.99} \\
			\hline
		\end{tabular}
	\end{center}
	\vskip -0.15in
\end{table}

\subsection{Deepgoing exploration} \label{sec:deepexplore}
\subsubsection{IPMC with different discrepancy metric}
To further explore the character of the distribution alignment, we conducted an ablation experiment employing different discrepancy metrics for the proposed approach, i.e., IPMC(\textit{Fp} + \textit{DA}) - KL (KL-Divergence) and IPMC(\textit{Fp} + \textit{DA}) - WD (Wasserstein Distance). See Tab. \ref{tab:e} for the results, which indicate that adopting either KL-divergence or Wasserstein distance as the discrepancy metric can enhance the performance of the proposed model, compared with the ablation model IPMC(\textit{Fp}). Yet the improvements of taking different discrepancy metrics are inconsistent, and accordingly IPMC(\textit{Fp} + \textit{DA}) - WD beats IPMC(\textit{Fp} + \textit{DA}) - KL with an advance by 1.72\% on CIFAR-10, and 0.82\% on CIFAR-100.

\subsubsection{Experiments under different settings of the multiple views} \label{sec:extendedcomp}
We conducted comparisons by grouping different views as our input for the proposed IPMC. As demonstrated in Tab. \ref{tab:f}, the results indicate that generally adopting more views as the input can enhance the performance of the proposed model. For details, \textit{IPMC w/ RGB-L-ab} outperforms other comparative methods. As we discussed in Sec. \ref{sec:method}, multiple views improve the method by restricting the learned representations with the added noisy information (different from the original input $X$). Therefore, we conjecture that if one view contains more different data, it is more possible for the model to learn a discriminative representation by adopting this view. This is supported by the experiment, for example, \textit{IPMC w/ RGB-L} improves \textit{IPMC w/ RGB} and \textit{IPMC w/ L} with a significant advance, etc. Furthermore, it is widely acknowledged that the RGB view has three channels (i.e., Red, Green, and Blue), the ab view has two channels (i.e., a and b), and the L view only has one channel. By observation, we found that \textit{IPMC w/ ab} beats \textit{IPMC w/ L} and \textit{IPMC w/ RGB} beats \textit{IPMC w/ ab}, which also proves our conjecture. Yet there is an exception that \textit{IPMC w/ RGB-L} beats \textit{IPMC w/ RGB-ab} with an advance by 1.31\% on Tiny ImageNet, and 0.76\% on STL-10, and our consideration lies in that the reason for the inconsistent improvements is that both RGB view and ab view describe the color-related information, while L view can depict the outline information of objects to some degree. So the information in RGB view and ab view might be overlapped, and L view is a valuable supplement to these views. We also found that even by taking two views (e.g., \textit{RGB-L}, \textit{RGB-ab}, or \textit{L-ab}) IPMC outperforms \textit{CMC w/ RGB-L-ab} on STL-10, and \textit{IPMC w/ RGB-L} beats \textit{CMC w/ RGB-L-ab} on Tiny ImageNet, which further validates the effectiveness of the proposed method.

\begin{figure*}
	\begin{center}
		\centerline{\includegraphics[width=1.8\columnwidth]{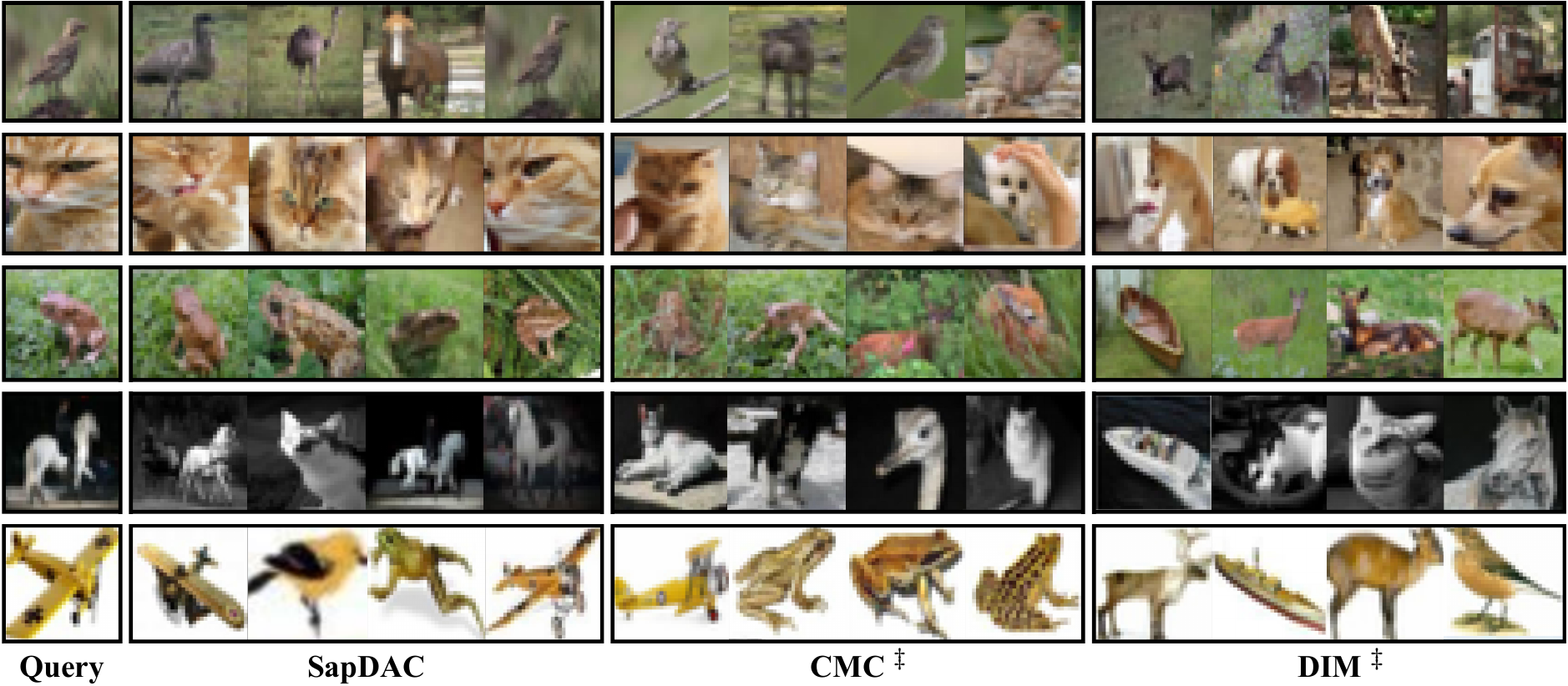}}
		\caption{Extended representation visual comparison for studying the merits of IPMC on the CIFAR-10 dataset. To evaluate the learned representations, we conducted the nearest-neighbor using $L_1$ distance to measure the discriminability of the representations. The leftmost images are randomly selected images from the CIFAR-10 dataset as queries, and the other images are their nearest neighbors measured in the representations of IPMC(\textit{Sap} + \textit{DA}), CMC, and DIM, respectively. We reimplemented CMC straightforward following the architecture proposed by the paper \cite{Tian2019Contrastive} and only adopt it on the CIFAR-10 dataset, and DIM is reimplemented by following the setting of \cite{hjelm2018learning}.}
		\label{fig:visualization}
	\end{center}
	\vspace{0.5cm}
\end{figure*}

\subsubsection{Hyper-parameter influences} \label{sec:parastudy}
For the sake of highlighting the impacts of hyper-parameters, we performed experiments with a slice of parameters used in the proposed method. The Tiny-ImageNet dataset is adopted for the parameter study experiments, since Tiny-ImageNet has various categories and a larger amount of examples, and consequently, the results derived from it are stable. The backbone encoder is conv network as in Tab. \ref{tab:a}.

Specifically, we performed several experiments to study the impacts of the hyper-parameters. The hyper-parameter $\beta$ balances the impact of the proposed pool contrastive representation learning approach and the representation distribution alignment. The hyper-parameters $\gamma$ and $\delta$ are proposed to leverage the loss of pool CL. For details, $\gamma$ is the parameter to balance the $log{}$ term, and $\delta$, as the interval factor between similarities, focuses on adjusting the interval of the positive similarity and the negative similarity. To intuitively understand the parameters' influences, we took experiments based on the classification task on Tiny ImageNet.

{As the results are manifested in Fig. \ref{fig:heatmap}, the plots further elaborate our parameter studies' results with IPMC(\textit{Sap} + \textit{DA}) on the benchmark dataset. To explore the influence of $\gamma$ and $\delta$, we first fixed $\beta$, and then we selected $\gamma$ from the range of \{$2^{2}, 2^{3}, 2^{4}, 2^{5}, 2^{6}, 2^{7}$\} and $\delta$ from the range of \{$0.20, 0.25, 0.30, 0.35, 0.40, 0.45$\}. Following the same experimental principle as above, we selected $\beta$ from the range of \{$10^{-4}, 10^{-2}, 1, 10^{2}, 10^{4}, 10^{6}$\}. See $(a)$, $(b)$, and $(c)$ shown in Fig. \ref{fig:heatmap} for the details of the comparison. It is observed that appropriate enhancement of feature discriminability can improve the performance of our proposed method. In general, good classification performance is highly dependent on the $\gamma$ term. Also, $\delta$ is an intensely necessary supplement for adapting the interval between similarities to enhance the similarity-measuring capacity of the model. As such, the potential to improve the learned representations grows with the adjustment of term $\beta$, and the feature distribution alignment helps IPMC in classification performance with a suitable $\beta$. A paramount reason behind the above is that it aligns the distributions of views in the latent space, which improves the capacity of the method to learn shared information from a multi-view context with the representation distribution constraint.}

\begin{figure}
	\centering
	\includegraphics[width=0.3\textwidth]{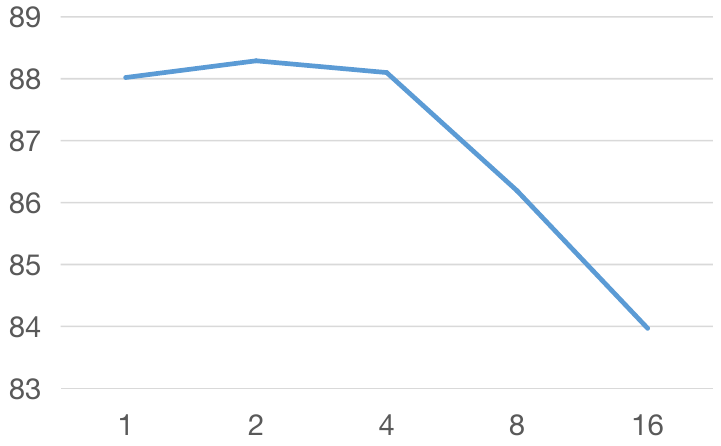}
	\caption{{Exploratory experiments of hyper-parameter k in the set-tier of IPMC.}}
	\label{fig:hpsapk}
	\vspace{0.5cm}
\end{figure}

\begin{figure}
	\centering
	\includegraphics[width=0.3\textwidth]{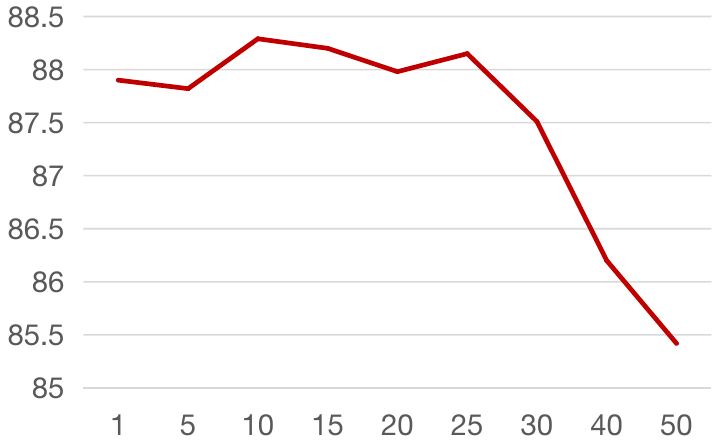}
	\caption{{Exploratory experiments of hyper-parameter $\eta$ in the set-tier of IPMC.}}
	\label{fig:hpsapeta}
	\vspace{0.5cm}
\end{figure}

{For the hyper-parameters in the self-adjusted pool module, we focus on exploring the impacts of k and $\eta$ on the model performance, since k and $\eta$ directly control the candidate negative samples for transferring. The exploratory experiments are conducted on the CIFAR10 dataset with the fc backbone encoder. As the results are manifested in Fig. \ref{fig:hpsapk} and Fig. \ref{fig:hpsapeta}, the plots demonstrate the influence curve of the hyper-parameters on the classification accuracy on downstream tasks. In detail, for the experiments of hyper-parameter k in Fig. \ref{fig:hpsapk}, relatively-less candidate negative samples for transferring fit the learning paradigm of IPMC, which maintain the consistency of the training process. The reason behind such a phenomenon is that as demonstrated in Fig. \ref{fig:cltrainplot}, at the beginning of contrastive learning, the negative samples and \textit{fake} negative samples are mixed together, such that the transferring of the proposed self-adjusted pools holds relatively low credibility. Due to the adopted moving-average mechanism, the low credibility can be accumulated, which in turn leads to the erroneous filtering of false negative samples. For the receptive field hyper-parameter $\eta$, as demonstrated in Figure \ref{fig:hpsapeta}, the over-small $\eta$ leads to the over-sensitive towards the current training step of the model, while as the aforementioned discussion, the over-large $\eta$ leads to the excessive emphasis on erroneous transferring in the early stage of training of the model, and the large $\eta$ requires the relatively large memory bank. Thus, the appropriate setting of $\eta$ can further promote the improvement of the SSL model. Additionally, based on the empirical observations, we conclude that the variations in values of k and $\eta$ have limited impacts on the time complexity, since compared with the transferring process of self-adjusted pools, the other processes of IPMC are relatively time-consuming, and compared with the main memory bank for features of candidate negative samples, the memory bank affected by $\eta$ has little impacts on the space complexity of the whole model.}

\subsection{{The robustness evaluation of IPMC towards data perturbations}} \label{sec:diffaug}
{To demonstrate the robustness of IPMC towards data perturbations, we conduct multiple comparisons on Tiny ImageNet by adopting the fc encoder, and the results are shown in Tab. \ref{tab:diffaug}. Note that most of the candidate data perturbations are similar to the data augmentations leveraged by the benchmark baselines \cite{chen2020simple, Tian2019Contrastive, 2020Kaiming, 2020Bootstrap}, including rotate, random crop, random grey, color jitter. For the random mask, we follow the benchmark masking approach of state-of-the-art masked image modeling methods \cite{DBLP:conf/cvpr/HeCXLDG22, DBLP:conf/iclr/Bao0PW22} while adopt the perturbation rate shared with the random grey. The intuition behind such a behavior is that the intrinsic self-supervised tasks between the masked image modeling methods and the contrastive methods are different, and then the information acquired by the representations learned by these methods are inconsistent, for instance, the masked image modeling methods focus on learning the image recovery information, while the contrastive methods are dedicated to model the discriminative information, such that directly adopting the perturbation rate of the masked image modeling methods \cite{DBLP:conf/cvpr/HeCXLDG22, DBLP:conf/iclr/Bao0PW22} leads to the representation collapse of contrastive methods. We adopt CMC and IPMC as the compared methods, and the reason is that the proposed IPMC is based on the benchmark baseline CMC, such that the head-to-head comparisons between these methods can significantly demonstrate the robustness superiority of IPMC over the baseline method towards data perturbations.}

{For the comparison results, we observe from Tab. \ref{tab:diffaug} and disclose that IPMC outperforms the compared method in all performed comparisons. It is worth noting that even using inappropriate data perturbations degenerates the performance of the proposed method and the benchmark method, but the performance of our method is still better than that of the compared method, e.g., from the first to the sixth comparisons, we find that the performance gaps between IPMC and CMC are preserved in the range of 0.57\% to 1,42\%. For the last two comparisons, we observe that the performance of IPMC and CMC is inconsistent, which is because of the incompatibility between the random mask and the paradigm of contrastive learning. Moreover, such a data perturbation is relatively function-overlapped with the random grey, such that although leveraging the random mask may improve the performance of IPMC on Tiny ImageNet, we still exclude such a data perturbation in the training on benchmarks. Concretely, the various comparisons using different combinations of data perturbations sufficiently prove the significant robustness of the proposed IPMC.}

\subsubsection{Representation visual comparison}
\begin{figure}
	\centering
	\includegraphics[width=0.4\textwidth]{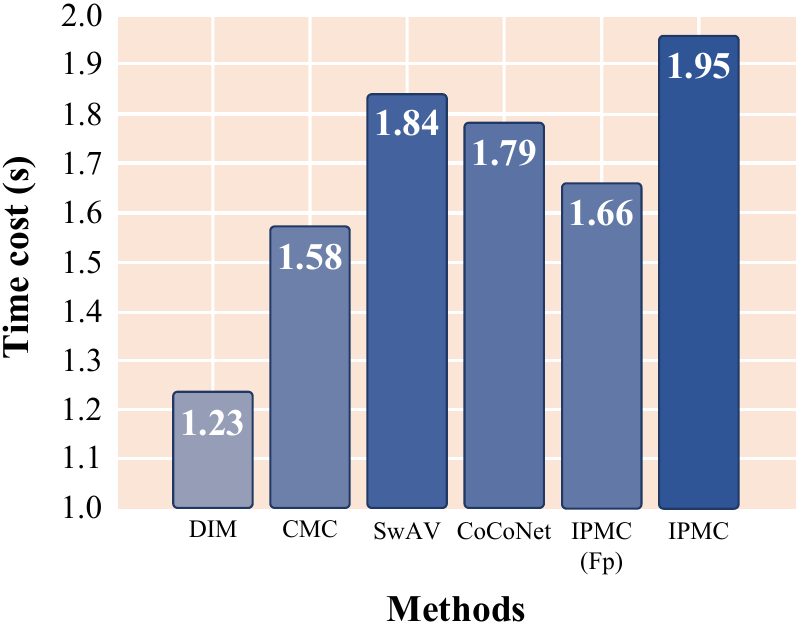}
	\caption{{The average computational time cost comparisons, which are performed based on the training of a batch. In detail, the candidate process includes the feed-forward calculation and the back-propagation training of the encoders.}}
	\label{fig:tcchart}
	\vspace{0.5cm}
\end{figure}

For the sake of clarifying the metric structure of IPMC’s representations, we conducted visual comparisons to explore the performance of the learned representations of IPMC(\textit{Sap} + \textit{DA}), CMC \cite{Tian2019Contrastive} and DIM \cite{hjelm2018learning}, which is based on nearest-neighbor of $L_1$ distance. Firstly, we randomly chose a sample from each class in the dataset and then sorted the images used for comparison in terms of the $L_1$ distance in the latent space by comparing the representations of all three methods together to avoid multiple occurrences of the same images. Lastly, the related 12 images (4 images for IPMC, 4 images for CMC, and the last 4 images for DIM) are selected with the lowest $L_1$ distance respectively. As demonstrated in Fig. \ref{fig:tcchart}, the representations learned by IPMC, which is the complete version, are more discriminative and have more structures that are easier interpreted, since neighboring representations correspond to visually similar images of the same category. There are several reasons behind this circumstance. First and foremost, the proposed IPMC learns multi-view representations that are more discriminative than single-view representations, which proves that the representations built by IPMC and CMC (also learning multi-view representations) have relatively significant improvement compared with DIM. Furthermore, our proposed anchor-free CL method and the unified loss jointly help IPMC to refine the representations by improving multi-view feature discriminability. Last, but not least, the inter-view representation aligning method enhances the learned representations' inter-view discriminability by considering the discrepancy metric. These improvements jointly strengthen the built representations of IPMC. The findings in the representation visual comparison confirm that benefiting from the proposed novel methodology, IPMC outperforms prior and current relevant approaches in the self-supervised representation learning research area.

\section{{Limitation discussions}}
\subsection{{Discussion on the time complexity}}
{In head-to-head comparisons, IPMC achieves the state-of-the-art, which demonstrates that the behaviors of IPMC in the three-tier progressive manner indeed enhance the model to learn discriminative information from the inputs. Yet compared with benchmark self-supervised methods, the time complexity of IPMC is relatively larger during training.}

{Specifically, as demonstrated in Fig. \ref{fig:tcchart}, we compare the time costs of the complete IPMC and the variation of IPMC, i.e., IPMC (Fp), with the benchmark methods, including DIM \cite{hjelm2018learning}, CMC \cite{Tian2019Contrastive}, SwAV \cite{2020Mathilde}, and CoCoNet \cite{li2022modeling}. By observing the comparison results, we find that due to the simple architecture and loss function, DIM achieves the lowest time costs during optimization in the head-to-head time complexity comparisons, and due to the complex view settings, SwAV has the highest time costs among the benchmark methods. For the proposed method, the time cost of IPMC is the highest, but the differences are not extremely significant, e.g., the time cost of IPMC is only higher than that of DIM by 0.72s, which is consistent with the optimization experiments in Sec. \ref{sec:opt}. Additionally, the time cost of IPMC (Fp) is even lower than that of SwAV and CoCoNet. The observation demonstrates that the view distribution alignment in the distribution-tier and the self-adjusted pool contrast in the set-tier indeed raise the time cost of the model during training, but according to the ablation comparison results on benchmarks, shown in Tab. \ref{tab:a} and Tab. \ref{tab:b}, such parts of IPMC can significantly improve the model's performance. During the test, due to the shared inference evaluation principle, the compared methods hold the same test time complexity. Concretely, with the empirical evidence, we state that IPMC outperforms the compared method (e.g., DIM, CMC, SwAV, CoCoNet) by significant margins, and the increase of the time cost is relatively limited.}

\subsection{{Threats to validity}}
{Following the validity threat analysis theory \cite{DBLP:books/daglib/0029933}, we explore the validity threats in a one-by-one manner.}

{For the conclusion validity, we follow the benchmark experimental settings \cite{hjelm2018learning, Tian2019Contrastive, chen2020simple}, e.g., choice of statistical tests, choice of sample size, etc. In order to avoid the threat to validity caused by imbalanced datasets, we perform multiple head-to-head experiments on various datasets, especially including a large-scale dataset ImageNet \cite{2009Feifei}, and the results are shown in Tab. \ref{tab:b}, such that the derived conclusion is validated.}

{For the internal validity, we impose sufficient ablation studies, and the corresponding discussions are introduced among the various experiments, e.g., the ablation setting and discussions in Sec. \ref{sec:ablresults}, which can prove the effectiveness of the proposed parts of IPMC. To further explore whether removing the proposed components of IPMC may affect the conclusion that \textit{``the proposed method leads to the improvement in model performance results''}, we conduct direct comparisons in Fig. \ref{fig:alphadeltastats} (a), and the results can support the effectiveness of the proposed components of IPMC.}

{For the construct validity, theoretically, multi-view SSL methods, including the proposed IPMC, share a foundational assumption, which is proved by benchmark analyses \cite{2022Chaos, 2008Sridharan, 2013Xu}. Empirically, visual multi-view SSL methods \cite{Tian2019Contrastive, chen2020simple, 2020Mathilde} demonstrate the applicability of such an assumption to image-related and video-related tasks, and the graph-based multi-view SSL method \cite{you2020graph} demonstrate the applicability of the multi-view assumption to graph-related tasks. Concretely, the assumption held by the proposed IPMC is theoretically and empirically proved.}

{For the external validity, to avoid the performance inconsistency caused by the random factors, e.g., random seeds, random data perturbations, etc., we collect the results of 20 trials for comparisons. The average result of the last 20 epochs is used as the final result of each trial. The average results from total of 20 trials are presented in tables, and the 95\% confidence intervals are also reported. Note that the results without 95\% confidence intervals are quoted from the original papers. Additionally, we conduct comparisons on multiple downstream tasks, including image classification tasks, graph prediction tasks, and action recognition tasks, to avoid the influence of artificial experimental settings on the experimental results. Concretely, according to the sufficient observations on the experimental results, we demonstrate that IPMC can consistently outperform benchmark self-supervised methods.}

\section{Conclusion} \label{sec:conclusion}
We rethink the self-supervised MVL from the perspective of information theory and then propose the information theoretical framework of generalized multi-view self-supervision. Guided by it, we develop a three-tier heuristic progressive method, called IPMC, to learn consistent and sufficient representations. IPMC performs the view alignment in the distribution tier, constructs the self-adjusted pool contrast in the set tier, and employs a unified loss in the instance tier. Intensive theoretical analyses and experimental comparisons manifest that IPMC achieves state-of-the-art.

\section*{Acknowledgements} \label{sec:conclusion}
We thank the reviewers for their efforts in revising this article. This work is supported by the 2022 Special Research Assistant Grant project (E3YD590101).

\bibliographystyle{ACM-Reference-Format}
\bibliography{reference}

\clearpage
\appendix

\section{Theoretical analyses}
In this section, we provide several theoretical proofs for the proposed theorems and remarkable deepgoing analyses.

\subsection{Proofs} \label{sec:proof}

{\bf{Proof of Theorem \ref{the:1}}} \label{pro:1}
To understand the improvement of the consistency constraint toward the self-supervision of the learned representation, we clarify the potential of optimizing $Y$ to achieve $Y^*$ by validating that there exists a $Y^*$ s.t. $H(Y^*) \leq H(Y)$, which is proved by introducing the KL-divergence \cite{1951Ls} measurement into the calculation of MI:

\begin{proof}
	To proof that there exists a $Y^*$ s.t. $H(Y^*) \leq H(Y)$\\
	\\Suppose $H(Y^*) = I(X;V_1;V_2) + I^{min}(Y;X|V_1;V_2) + I^{min}(Y;V_1|X;V_2) + I^{min}(Y;V_2|X;V_1)$\\
	\\$H(Y) = I(X;V_1;V_2) + I(Y;X|V_1;V_2) + I(Y;V_1|X;V_2) + I(Y;V_2|X;V_1)$\\
	\\Therefore, $Y^*$ can make $I(X;V_1;V_2) + I^{min}(Y;X|V_1;V_2) + I^{min}(Y;V_1|X;V_2) + I^{min}(Y;V_2|X;V_1) \leq I(X;V_1;V_2) + I(Y;X|V_1;V_2) + I(Y;V_1|X;V_2) + I(Y;V_2|X;V_1)$ hold\\
	\\$\because I(Y;X|V_1;V_2) = I(Y;X) - I(Y;X;V_1) - I(Y;X;V_2)$\\
	\\$\because H(X)$, $H(V_1)$, and $H(V_2)$ are constant\\
	\\$\therefore$ $I^{min}(Y;X) + I^{min}(Y;V_1) + I^{min}(Y;V_2) \leq I(Y;X) + I(Y;V_1) + I(Y;V_2)$\\
	\\$\because I(Y;X) = \sum\limits_{y \in Y}\sum\limits_{x \in X}{\mathcal{P}(y, x)}{\log\frac{\mathcal{P}(y, x)}{{\mathcal{P}(y)\cdot}\mathcal{P}(x)}}$\\
	\\$\therefore I^{min}(Y;X) = \sum\limits_{y \in Y^*}\sum\limits_{x \in X}{\mathcal{P}(y, x)}{\log\frac{\mathcal{P}(y, x)}{{\mathcal{P}(y)\cdot}\mathcal{P}(x)}}\\= I(Y^*;X)$\\
	\\$\because I^{min}(Y;X) \leq I(Y;X)$\\
	\\$\therefore I(Y^*;X) \leq I(Y;X)$\\
	\\Since, $V_1$ and $V_2$ are two generated views of $X$, and they both have been deleted a part of $X$'s information, and the view-specific information have been added into $V_1$ and $V_2$ so that the sufficiency of self-supervision degenerates because of $\epsilon^{info}_i$ (proposed in Sec. \ref{sec:method}) that exists in view-specific information of $X$, $V_1$, or $V_2$. Therefore, compared with the compact representation $Y^*$ learned from the aligned views, the representation $Y$ learned from the unaligned views contains a certain $\delta^{info}$ so that the assumption of $\delta^{info}$ holds. In other words, there is a $\delta^{info}$ between $Y$ and $Y^*$, i.e., $Y^* = Y - \delta^{info}$.\\
	\\$\because I(Y^*;X) = I(Y - \delta^{info};X)$\\
	\\$\therefore$ to prove the existence of $Y^*$, we only need to prove:\\
	\\$I(Y - \delta^{info};X) \leq I(Y;X)$\\
	\\KL-divergence is defined as:\\
	\\$D_{KL}(P||Q) = \int\limits{\mathcal{P}(x)\log\frac{\mathcal{P}(x)}{\mathcal{Q}(x)}}dx$\\
	\\The discrete form of KL-divergence is:\\
	\\$D_{KL}(P||Q) = \sum\limits{\mathcal{P}(x)\log\frac{\mathcal{P}(x)}{\mathcal{Q}(x)}}$\\
	\\We try to use KL divergence to fit the calculation of mutual information, and the $\mathcal{P}$ and $\mathcal{Q}$ are approximated by:\\
	\\$\hat{\mathcal{P}}(x) = \mathcal{P}(x, y)$\\
	\\$\hat{\mathcal{Q}}(x) = \mathcal{P}(x) \cdot \mathcal{P}(y)$\\
	\\Put $\hat{\mathcal{P}}(x)$ and $\hat{\mathcal{Q}}{(x)}$ into the above formula of the discrete KL-divergence:\\
	\\$D_{KL}(\mathcal{P}_{XY}||\mathcal{P}_X\mathcal{P}_Y) = \sum\limits_{x \in X}\sum\limits_{y \in Y}{\mathcal{P}(x, y)}{\log\frac{\mathcal{P}(x, y)}{{\mathcal{P}(x)\cdot}\mathcal{P}(y)}}$\\
	\\Then, we get:\\
	\\$D_{KL}(\mathcal{P}_{XY}||\mathcal{P}_X\mathcal{P}_Y) = I(X;Y)$\\
	\\$\therefore I(Y;X) = D_{KL}(\mathcal{P}_{YX}||\mathcal{P}_{Y}\mathcal{P}_X)$\\
	\\$\therefore I(Y - \delta^{info};X) = D_{KL}(\mathcal{P}_{(Y - \delta^{info})X}||\mathcal{P}_{Y - \delta^{info}}\mathcal{P}_X)$\\
	\\Because $Y$ is not fully compact, which means $\delta^{info} \geq 0$. For the KL-divergence, $\mathcal{P}_{X}$ is constant, and $Y \geq \{Y-\delta^{info}\}$. Therefore, compared with the joint $\mathcal{P}_{YX}$ and $\mathcal{P}_{Y}\mathcal{P}_X$, the distributions of the joint $\mathcal{P}_{(Y - \delta^{info})X}$ and $\mathcal{P}_{Y - \delta^{info}}\mathcal{P}_X$ are more consistent, and then we get:\\
	\\$I(Y - \delta^{info};X) \leq I(Y;X)$\\
	\\$\therefore Y - \delta^{info}$ makes $I^{min}(Y;X|V_1;V_2) \leq I(Y;X|V_1;V_2)$ hold\\
	\\and therefore, $I(X;V_1;V_2) + I^{min}(Y;X|V_1;V_2) + I^{min}(Y;V_1|X;V_2) + I^{min}(Y;V_2|X;V_1) \leq I(X;V_1;V_2) + I(Y;X|V_1;V_2) + I(Y;V_1|X;V_2) + I(Y;V_2|X;V_1)$ holds\\
	\\$\therefore$ there exists a $Y^*$ s.t. $H(Y^*) \leq H(Y)$. Specifically, $Y^* = Y - \delta^{info}$
\end{proof}
\begin{figure}[t]
    \centering
    \includegraphics[width=0.45\textwidth]{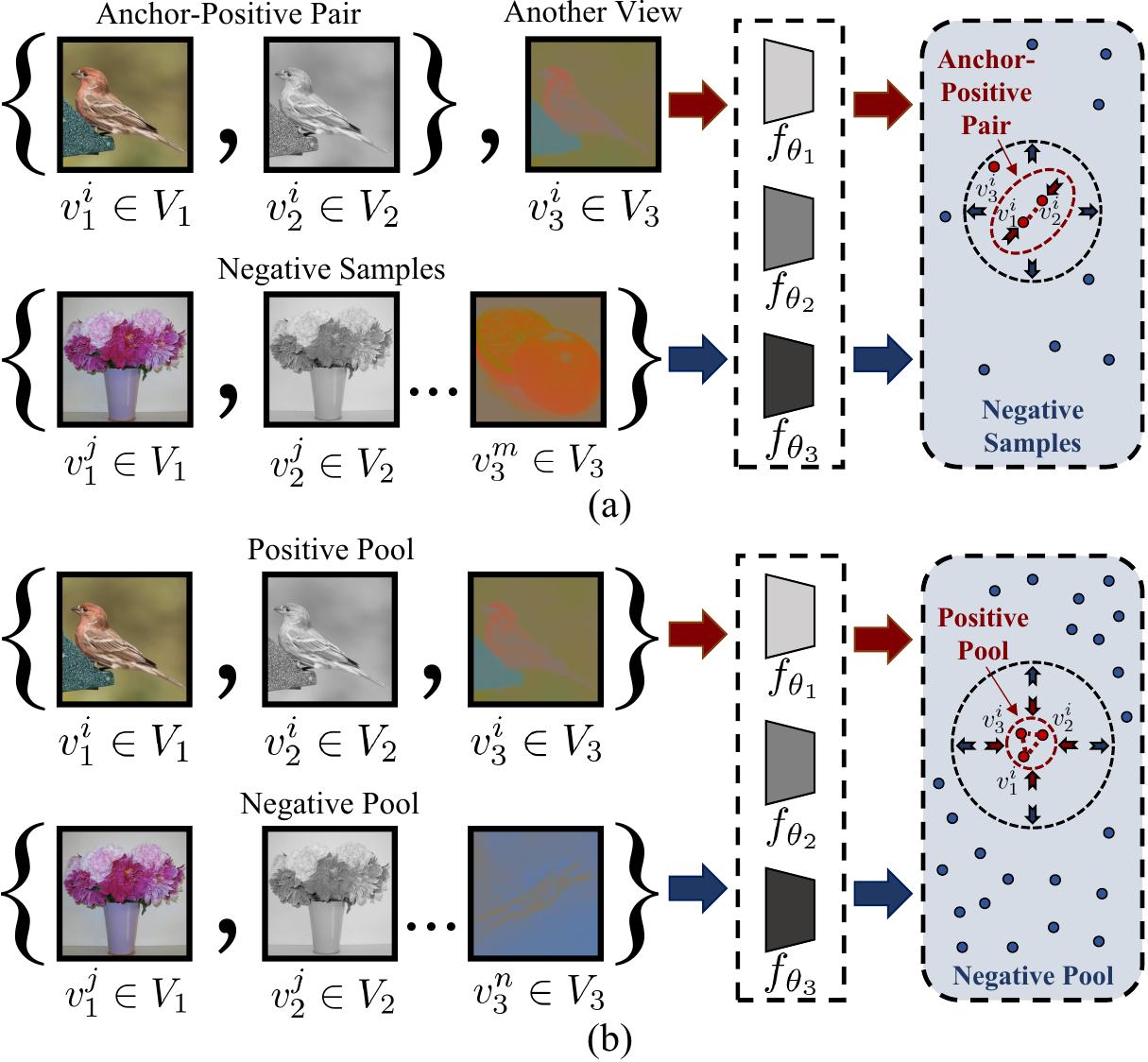}
    \caption{Comparison between the vanilla anchor-based contrastive learning framework (a) and our proposed pool contrastive learning framework (b).}
    \label{fig:pdd}
    \vspace{0.5cm}
\end{figure}
\subsection{Remarks on the difference of conventional contrast and the proposed pool contrast} \label{sec:remarks}
As demonstrated in Fig. \ref{fig:pdd}, we demonstrate an example of learning representations from three views: optical Red-Green-Blue (RGB) view ($V_1$), the luminance channel (L) view ($V_2$), and the ab-color channel (ab) view ($V_3$). In (b), only the initial fixed-pool version of our IPMC is shown. We further put dynamically self-adjusted pool into practice on subsequent training, which is shown in Fig. \ref{fig:algoframe}. Therefore, our model can jointly involve more positive terms (including the selected views from other samples) and negative terms in CL.

\subsection{Algorithm description} \label{sec:algocode}
In this paper, we introduce a novel unsupervised representation learning approach, i.e., \textit{Information theory-guided heuristic Progressive Multi-view Coding} (IPMC), of which Fig. \ref{fig:alignment} and \ref{fig:algoframe} depict the overview framework. The following subsections provide the IPMC design details.

As shown in the following algorithm, our proposed IPMC is an end-to-end representation learning approach. We first build the unified loss of self-adjusted pool contrastive learning $\mathcal{L}_{UniSap}$ including two hyper-parameters: $\gamma$ and $\delta$, where $\alpha$ is replaced by $\delta$ as $O^{pos}=1+\delta$, $O^{neg}=-\delta$, $\delta^{pos}=1-\delta$, and $\delta^{neg}=\delta$ in Eq. \ref{eq:Luni}. Then, the loss of view distribution alignment $\mathcal{L}_{DA}$ is built with a hyper-parameter $\beta$ to balance the impacts between $\mathcal{L}_{UniSap}$ and $\mathcal{L}_{DA}$. The derived loss, namely $\mathcal{L}_{IPMC}$, is used in the back-propagation training process based on Adam gradient optimization.

The proposed IPMC is a generalized self-supervised representation learning approach designed for general application use for various downstream tasks, e.g., classification, clustering, regression, etc. We can directly attach the downstream tasks with IPMC and train them at the same time based on the training process of the end-to-end learning.

Here, we provide a pseudo-code for IPMC training loop in using PyTorch machine learning python library without the inclusion of the detailed matrix processing or helper utility functions \& codes that are irrelevant to the algorithm:

\begin{table*}[hb]
	\begin{python}
		# index: the index of samples in memory bank
		# model: view-wise backbone encoders, approximated by conv or fc
		# contrast: similarity calculation with using memory bank
		# critic: critic network (MLP) for Wasserstein distance calculation
		for x, index in loader: # load a batch x
    		l, ab, ori = model(x) # view-wise backbone encoding
    		# self-adjusted pool contrastive learning
    		# achieve similarities
    		out_ab2l,...,out_ori2ab = contrast(l, ab, ori, index)
    		# calculate unified loss
    		loss = criterion_gh(out_ab2l,...,out_ori2ab)
    		# view distributions alignment based on Wasserstein distance
    		wd_loss = calc_wd(critic, critic_optim, l, ab, ori)
    		loss += wd_loss
    		# Adam update
    		loss.backward()
    		optimizer.step()
		
		# unified loss calculation
		def criterion_gh(out_ab2l,...,out_ori2ab):
    		# split the similarities and clone the similarity set
    		pos_ab2l, neg_ab2l = torch.split(out_ab2l,[1,out_ab2l.shape[1]-1],dim=1)
    		out_ab2l_cl = out_ab2l.squeeze(-1).T[1:].T.unsqueeze(-1).clone()
    		...
    		# self-adjusted pool process
    		# topK: top k nearest neighbors (hyper-parameter)
    		for _ in topK: # usually topK = 1
        		# pick out most similar fake negative terms
        		max_ab2l_values, max_ab2l_pos = torch.max(out_ab2l_cl, dim=1)
        		...
        		# transfer to positive pool
        		pos_ab2l = torch.cat((pos_ab2l, max_ab2l_values), dim=1)
        		neg_ab2l = del_moved_ele(neg_ab2l, max_ab2l_pos)
        		out_ab2l_cl = del_moved_ele(out_ab2l_cl, max_ab2l_pos)
        		...
    		# calculate the unified loss
    		pos = torch.cat((pos_ab2l, pos_l2ab, pos_ori2l, pos_l2ori, pos_ab2ori, pos_ori2ab), dim=1)
    		neg = torch.cat((neg_ab2l, neg_l2ab, neg_ori2l, neg_l2ori, neg_ab2ori, neg_ori2ab), dim=1)
    		# alpha_p, alpha_n, delta_p, delta_n, gamma: hyper-parameters
    		logit_p = - alpha_p * (pos - delta_p) * gamma
    		logit_n = alpha_n * (neg - delta_n) * gamma
    		loss = soft_plus(torch.logsumexp(logit_n, dim=1) + torch.logsumexp(logit_p, dim=1)).sum().div(pos.shape[0])
    		loss = loss.div(gamma).mul(16).sum()
    		return loss
		
		# train critic and calculate total Wasserstein distance
		def calc_wd(critic, critic_optim, l, ab, ori):
    		# k_critic, hypergp, beta: hyper-parameters
    		for a, b in [l, ab, ori] and not a == b:
        		for _ in range(k_critic):
            		# calculate the gradient penalty of critic network
            		gp = gradient_penalty(critic, a, b)
            		wasserstein_distance = critic(a).mean() - critic(b).mean()
            		critic_cost = -wasserstein_distance + hypergp * gp
            		critic_cost.backward()
            		critic_optim.step()
            		# calculate Wasserstein distance
            		set_requires_grad(critic, requires_grad=False)
            		wasserstein_distance = (critic(l).mean() - critic(ab).mean())
            		wasserstein_distance += (critic(l).mean() - critic(ori).mean())
            		wasserstein_distance += (critic(ab).mean() - critic(ori).mean())
    		wd_loss = beta * wasserstein_distance
    		return wd_loss
	\end{python}
\end{table*}

\end{document}